\documentclass[preprints,article,accept,moreauthors,pdftex]{Definitions/mdpi}

\usepackage{todonotes}
\DeclareMathOperator*{\argmin}{arg\,min}
\usepackage{caption}
\usepackage{subcaption}
\usepackage{makecell}

\usepackage{algorithm}
\usepackage{algpseudocode}

\firstpage{1}
\pubvolume{1}
\issuenum{1}
\articlenumber{0}
\pubyear{2022}
\copyrightyear{2022}
\datereceived{}
\dateaccepted{}
\datepublished{}
\hreflink{https://doi.org/} 

\Title{Defending against Reconstruction Attacks through Differentially Private Federated Learning for Classification of Heterogeneous Chest X-Ray Data}

\TitleCitation{Defending against Reconstruction Attacks through Differentially Private Federated Learning for Classification of Heterogeneous Chest X-Ray Data}

\Author{Joceline Ziegler $^{1,2,}$*\orcidA{}, Bjarne Pfitzner $^{1,2}$\orcidB{}, Heinrich Schulz $^{3}$\orcidC{}, Axel Saalbach $^{3}$\orcidD{} and Bert Arnrich $^{1,2}$\orcidE{}}

\AuthorNames{Joceline Ziegler, Bjarne Pfitzner, Heinrich Schulz, Axel Saalbach and Bert Arnrich}

\AuthorCitation{Ziegler, J.; Pfitzner, B.; Schulz, H.; Saalbach, A.; Arnrich, B.}

\address{%
$^{1}$ \quad University of Potsdam, Digital Engineering Faculty, 14482 Potsdam, Germany\\
$^{2}$ \quad Hasso Plattner Institute for Digital Engineering gGmbH, 14482 Potsdam, Germany; \{firstname\}.\{lastname\}@hpi.uni-potsdam.de\\
$^{3}$ \quad Philips Research, 22335 Hamburg, Germany; \{firstname\}.\{lastname\}@philips.com}

\corres{Correspondence: post@jocelineziegler.de
}

\abstract{Privacy regulations and the physical distribution of heterogeneous data are often primary concerns for the development of deep learning models in a medical context. This paper evaluates the feasibility of differentially private federated learning for chest X-ray classification as a defense against data privacy attacks. To the best of our knowledge, we are the first to directly compare the impact of differentially private training on two different neural network architectures, DenseNet121 and ResNet50. Extending the federated learning environments previously analyzed in terms of privacy, we simulated a heterogeneous and imbalanced federated setting by distributing images from the public CheXpert and Mendeley chest X-ray datasets unevenly among 36 clients. Both non-private baseline models achieved an area under the receiver operating characteristic curve (AUC) of $0.94$ on the binary classification task of detecting the presence of a medical finding. We demonstrate that both model architectures are vulnerable to privacy violation by applying image reconstruction attacks to local model updates from individual clients. The attack was particularly successful during later training stages. To mitigate the risk of privacy breach, we integrated Rényi differential privacy with a Gaussian noise mechanism into local model training. We evaluate model performance and attack vulnerability for privacy budgets $\varepsilon \in \{1,3,6,10\}$. The DenseNet121 achieved the best utility-privacy trade-off with an AUC of $0.94$ for $\varepsilon=6$. Model performance deteriorated slightly for individual clients compared to the non-private baseline. The ResNet50 only reached an AUC of $0.76$ in the same privacy setting. Its performance was inferior to that of the DenseNet121 for all considered privacy constraints, suggesting that the DenseNet121 architecture is more robust to differentially private training.
}


\begin{document}


\section{Introduction}

The development of machine learning models for medical use cases often requires collecting large amounts of sensitive patient data. Medical datasets are usually scattered across multiple sites and underlie rigorous privacy constraints of both ethical and regulatory nature \cite{rieke2020FutureDigital}. The effectiveness of anonymization to enable data sharing is dependent on the type of data and cannot always prevent re-identification \cite{kaissis2020SecurePrivacypreserving}. Federated learning gains increasing attention as a method for training machine learning models on distributed data in a privacy-preserving manner. In a federated learning setting, holders of sensitive data can make their data available for machine learning without sharing it with other parties. In several iterations, a central server distributes an initial model to several clients holding the data, e.g., medical institutions, which then individually train their models and provide them to the server for aggregation.

Federated learning provides a basic level of privacy by the principle of data minimization, i.e., data collection and processing are restricted to a necessary minimum. However, it cannot by itself formally guarantee privacy \cite{kairouz2021AdvancesOpen}. It has been shown that input data can successfully be reconstructed from model gradients \cite{geiping2020InvertingGradients, zhu2019DeepLeakage}. In addition to the threat of data reconstruction, attacks disclosing the presence of a specific data sample or property in the training data imply a serious privacy risk for individual contributors~\cite{naseri2021RobustnessPrivacy}.

Measures to prevent privacy breaches of machine learning models are subject to ongoing research. Differential privacy is a concept actively explored in this field. Intuitively, the goal of differential privacy is to limit the impact of a single data sample or a subset of the data on the outcome of a function computed on the data, thereby providing a guarantee that no or little information can be inferred about individual samples~\cite{naseri2021RobustnessPrivacy}. However, the application of differential privacy is known to decrease the utility of the machine learning model, characterized by a trade-off between utility and privacy specific to each use case~\cite{abadi2016DeepLearning,kairouz2021AdvancesOpen,kaissis2021EndtoendPrivacy,li2019PrivacyPreservingFederated}. Despite the potential of differentially private federated learning in healthcare, there has been an increased research interest only recently on selected use cases.

Vast amounts of medical image data are currently produced in daily medical practice. Chest X-rays play an essential role in diagnosing a variety of diseases such as pneumonia~\cite{rajpurkar2017CheXNetRadiologistLevel}, and also recently in studying COVID-19~\cite{feki2021FederatedLearning}. Automatic diagnosis assistance may substantially support the work of radiologists, which is particularly of interest in the face of ongoing medical specialist shortages~\cite{rimmer2017RadiologistShortage}. Digital support systems may also mitigate the impact of error sources in human assessment that occur systematically, e.g., due to increased workload and varying professional experience~\cite{itri2018FundamentalsDiagnostic}. Increased costs for the healthcare system and potentially fatal misdiagnoses can thereby be avoided.

We evaluate the potential of privacy-preserving federated learning for the use case of disease classification on chest X-ray images. As a key contribution, we directly compare two popular image classification model architectures, DenseNet121 and ResNet50, in terms of the effects of differentially private training on model performance and privacy preservation. Extending previous work, we introduce a federated environment that is subject to data heterogeneity and imbalance. We demonstrate that the basic federated learning setting is vulnerable to privacy violation through the successful application of reconstruction attacks. We specifically compare the vulnerability to privacy breach and the effect of differential privacy on a previously unconsidered complex model, DenseNet121, with the previously studied ResNet architecture. Our results endorse the conjecture that reconstruction attacks pose a realistic threat within the federated learning paradigm, even for large and complex model architectures. We integrate Rényi differential privacy into the federated learning process and investigate how it affects the utility-privacy trade-off for our use case. Two measures of privacy are addressed: The privacy budget $\varepsilon$ as part of the formal differential privacy guarantee and the susceptibility of the local models to reconstruction attacks. Our results suggest that the DenseNet121 is a promising architecture for feasible privacy-preserving model training on X-ray images. This novel insight may direct future research and applications in that area.

This paper is structured as follows: In Section~\ref{sec:related_work}, we briefly present previous work related to privacy-preserving federated learning for the task of X-ray classification. We introduce the used datasets in Section~\ref{sec:data} and explain our federated learning setup in Section~\ref{sec:FL}. We provide background information on the \textit{Deep Leakage from Gradients} attack (Section~\ref{sec:attack}) and on the integration of differential privacy into the training of neural networks (Section~\ref{sec:background_dp}). We present the results on model performance in a basic federated learning setting (Section~\ref{sec:fl_results}), demonstrate the susceptibility of our federated learning models to reconstruction attacks (Section~\ref{sec:results_attack}), and finally evaluate the impact of differential privacy on model performance and attack vulnerability (Section~\ref{sec:results_dp}). We discuss and summarize our findings in Sections~\ref{sec:discussion} and~\ref{sec:conclusion}.

\section{Related Work}
\label{sec:related_work}

The healthcare sector especially profits from privacy-preserving machine learning due to the natural sensitivity of the underlying patient data~\cite{qayyum2021SecureRobust, rieke2020FutureDigital, shah2021MaintainingPrivacy}. A wide range of applications demonstrate that federated learning is a potential fit for leveraging diverse types of medical data, including electronic health records~\cite{brisimi2018FederatedLearning}, genomic data~\cite{li2016VERTIcalGrid}, and time-series data from wearables~\cite{chen2020FedHealthFederated}. Examples related to medical image classification include brain tumor segmentation~\cite{li2019PrivacyPreservingFederated, sheller2019MultiinstitutionalDeep}, classification and survival prediction on whole slide images in pathology~\cite{lu2020FederatedLearning}, classification of functional magnetic resonance images (fMRI)~\cite{li2020MultisiteFMRI}, and breast density classification from mammographic images~\cite{roth2020FederatedLearning}. One large research area is concerned with the classification of chest X-ray images. \citet{calli2021deep} provided an overview of recent deep learning advances in this field, but do not consider federated learning. The feasibility of federated learning on chest X-rays has previously been benchmarked for both the CheXpert~\cite{irvin2019CheXpertLarge} and the Mendeley~\cite{kermany2018LabeledOptical} dataset. \citet{chakravarty2021FederatedLearning} enhance a ResNet18 architecture with a graph neural network for federated learning on CheXpert data with site-specific data distributions. \citet{nath2020EmpiricalEvaluation} deploy a DenseNet121 model for a real-world, physically distributed implementation of federated learning on CheXpert. \citet{banerjee2020MultidiseasesClassification} determine the ResNet18 architecture as superior for federated learning on Mendeley data in comparison with ResNet50, DenseNet121, and MobileNet.

\begin{specialtable}[h]
\centering
\caption{Overview of related works evaluating deep neural networks on the CheXpert or Mendeley datasets using DenseNet or ResNet architectures. The mentioned models are not necessarily exhaustive, some papers evaluate more ResNet and DenseNet architectures. We also include our paper at the bottom for comparison with related work.\\ \emph{(non-)IID} corresponds to a (not) independent and identical data distribution.\\
\emph{DP} corresponds to the use of differential privacy ($\varepsilon=6$.)}
\label{table:sc1_chexpert_comp}

\begin{minipage}{\textwidth}
\begin{tabular}{c l l l l}

\toprule
\textbf{Data} & \textbf{Reference} & \textbf{Model} & \textbf{Federated Learning} & \textbf{AUC}\\

\midrule
\multirow{7}{*}{\rotatebox[origin=c]{90}{CheXpert}} & \citet{irvin2019CheXpertLarge} & DenseNet121 & no & $0.889$ \\[0.25em]
& \multirow{2}{*}{\citet{bressem2020ComparingDifferent}} & DenseNet121 & \multirow{2}{*}{no} & $0.869$ \\
&  & ResNet50 &  & $0.881$ \\[0.25em]
& \multirow{2}{*}{\citet{ke2021CheXtransferPerformance}} & DenseNet121 & \multirow{2}{*}{no} & $0.859$\\
& & ResNet50 & & $0.859$\\[0.25em]
& \citet{chakravarty2021FederatedLearning} & ResNet18 & 5 sites, non-IID & $0.782$\\[0.25em]
& \citet{nath2020EmpiricalEvaluation} & DenseNet121 & 5 sites, IID & $0.803$\\[0.25em]
\midrule
\multirow{4}{*}{\rotatebox[origin=c]{90}{Mendeley}} & \citet{banerjee2020MultidiseasesClassification} & ResNet50 & 3 sites, non-IID \footnote{The data distribution between the three hospitals is $30:32:38$, with slightly varying class distributions.} & $0.976$\footnote{No AUC given, the value corresponds to the binary accuracy.}\\[0.25em]
& \multirow{3}{*}{\citet{kaissis2021EndtoendPrivacy}} & \multirow{3}{*}{ResNet18} & no & $0.93$ \\
& & & 3 sites\footnote{No information on the data distribution available.} & $0.92$ \\
& & & 3 sites$^c$, DP & $0.89$ \\
\midrule
\multirow{4}{*}{\rotatebox[origin=c]{90}{Both}} & \multirow{4}{*}{This paper} & \multirow{2}{*}{DenseNet121} & 36 sites, non-IID & $0.935$\\
& & & 36 sites, non-IID, DP & $0.937$ \\
& & \multirow{2}{*}{ResNet50} & 36 sites, non-IID & $0.938$\\
& & & 36 sites, non-IID, DP & $0.764$ \\
\bottomrule
\end{tabular}\\[0.25em]
\end{minipage}
\end{specialtable}

Surveys on current developments in the field of privacy-preserving machine learning and federated learning describe potential threat models and privacy attacks~\cite{enthoven2020OverviewFederated, kairouz2021AdvancesOpen, lyu2020ThreatsFederated}. \citet{zhu2019DeepLeakage} originally proposed the \textit{Deep Leakage from Gradients} (DLG) attack, which allows a malicious server instance to reconstruct complete data samples from received model gradients. Subsequent improvements of the idea include analytical label reconstruction~\cite{zhao2020IDLGImproved}, improved loss functions for gradient matching~\cite{geiping2020InvertingGradients, wang2020SAPAGSelfAdaptive} and an extension towards larger batch sizes~\cite{yin2021SeeGradients}. DLG and other privacy attacks have been identified as a severe threat to federated learning. \citet{wei2020FrameworkEvaluating} evaluate the impact of attack initialization, optimization method, and training parameters including batch size, image resolution, and activation function on DLG attack success for a small network.

Ensuring privacy and protecting against reconstruction in a practicable manner is not yet fully explored and remains an open problem for the federated learning paradigm~\cite{kairouz2021AdvancesOpen, li2020FederatedLearning, pfitzner2021FederatedLearning, rieke2020FutureDigital}. A key method that finds wide use among federated learning research is \textit{differential privacy}, first proposed by \citet{dwork2006DifferentialPrivacy} in the context of database systems. Generally, it describes the addition of carefully crafted noise into a system to prevent learning too much about single data instances and measuring the remaining risk. \citet{mironov2017RenyiDifferential} introduced the variant of \textit{Rényi differential privacy}, defining a tighter bound on the privacy loss. Differential privacy in federated learning is often achieved using \textit{differentially-private stochastic gradient descent} (DP-SGD)~\cite{abadi2016DeepLearning, li2020SecureFederated, truex2020LDPFedFederated}, an algorithm that determines the appropriate noise scale and how to clip the model parameter. The combination of federated learning and differential privacy has been explored in multiple medical use cases including prediction of mortality and adverse drug reactions from electronic health records~\cite{choudhury2020DifferentialPrivacyenabled}, brain tumor segmentation~\cite{li2019PrivacyPreservingFederated}, classification of pathology whole slide images~\cite{lu2020FederatedLearning}, detection of diabetic retinopathy in images of the retina~\cite{malekzadeh2021DopamineDifferentially}, and identification of lung cancer in histopathologic images~\cite{adnan2021FederatedLearning}.

Most similarly to this work, \citet{kaissis2021EndtoendPrivacy} demonstrate a framework for the implementation and evaluation of privacy-preserving machine learning in a federated learning setting on the Mendeley dataset evenly distributed among three clients. They combine a ResNet18 model with a secure multi-party computation protocol and differential privacy and compare the success of reconstruction attacks on centralized and federated learning models. We extend their setting by considering larger networks, simulating a scenario with heterogeneous data unevenly distributed among a larger number of clients, and evaluating the impact of different parameters on model performance and the model's vulnerability to reconstruction attacks.

\section{Materials and Methods}

In this section, we first provide information about the used datasets. Then, we go over the three central pieces of our article: the federated learning baseline and our heterogeneous data distribution, the reconstruction attack, and finally the introduction of differential privacy as a defence against the attack.

\subsection{Data}
\label{sec:data}

CheXpert \cite{irvin2019CheXpertLarge} comprises 224,316 images of 65,240 adult patients in total, where 234 images are labeled by professional radiologists for use as a validation set. We only considered frontal view images as this accounts for the higher prevalence of frontal view images in the clinical setting and ensures compatibility with the Mendeley dataset. Each image is labeled with one or more of 13 classes referring to a medically relevant finding, or with "No Finding". Following previous work \cite{lenga2020ContinualLearning, mitra2020SystematicSearch}, uncertain labels were considered as negative (U-Zeroes method).

The Mendeley chest X-ray dataset version 3 \cite{kermany2018LabeledOptical} contains 5,856 images of pediatric patients and is split into original training and test sets with 5,232 and 624 images. Each image is labeled as either "Normal", with "Viral Pneumonia", or "Bacterial Pneumonia". For convenience, we assume that "Normal" in the Mendeley dataset corresponds to "No Finding" in the CheXpert dataset. To ensure compatibility between the dataset labels, our primary setting is a binary classification task based on the "No Finding" or "Normal" label, indicating the presence or absence of a medically relevant condition.

\subsection{Federated Learning}
\label{sec:FL}

\begin{figure}
  \centering
  \includegraphics[width=\linewidth]{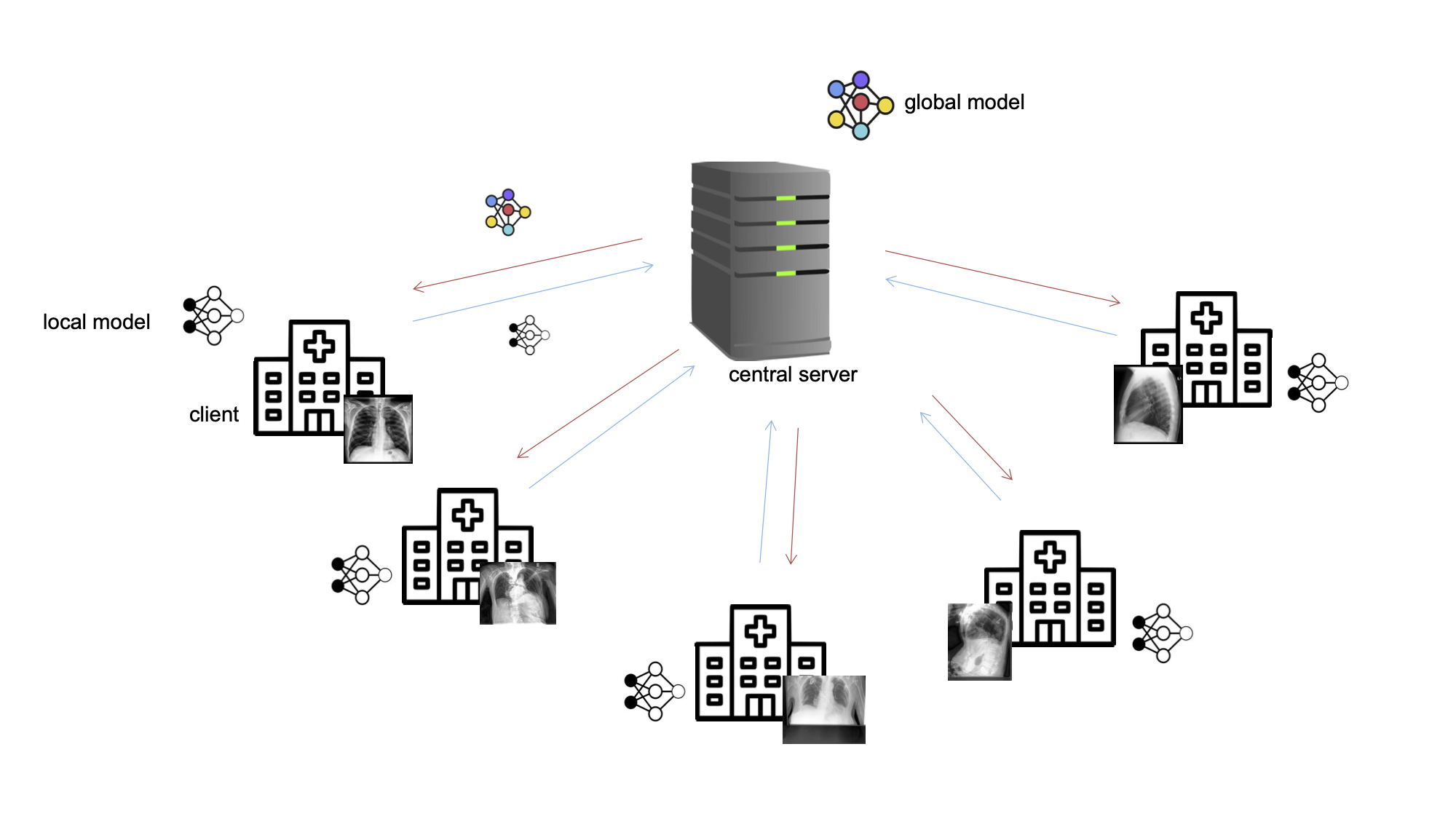}
  \caption{In the federated learning setup, the server first initializes a model and distributes the model parameters to its clients. Over several iterations, each client trains the model individually on its data for a defined number of local epochs, sends the parameters of its locally trained model back to the server for aggregation, and receives a global model, aggregated from all trained local models.}
  \label{fig:fl_overview}
\end{figure}

Successful training of a deep learning model usually relies on the availability of a single large, high-quality dataset, requiring prior data collection and curation, potentially associated with great expense in time and resources. Despite such efforts, data transfer or direct access to the data can still often not be granted due to patient privacy concerns. Federated learning enables model training on scattered data that remains at the participants' sites at all times \cite{mcmahan2017CommunicationEfficientLearning}.

A typical federated learning system consists of a central server that orchestrates the training procedure, and several clients that communicate with the server (Fig. \ref{fig:fl_overview}). The server initializes a model and distributes the model parameters to its clients. In parallel, a subset of clients trains the model individually on their data for a defined number of epochs, which is equivalent to local stochastic gradient descent (SGD) optimization. The clients send their local models back to the server, where they are aggregated through \textit{federated averaging} \cite{mcmahan2017CommunicationEfficientLearning}. The new global model is again distributed among the clients, and the process is repeated until convergence or until a defined number of communication rounds has been reached.

\subsubsection{Experimental Setup}
In real-world use cases, the expectation is that datasets between clients in a federated learning setting show some variety. We reflect this in our simulation of a federated environment by combining two public X-ray datasets representing heterogeneous target populations, adult and pediatric patients.
Our federated learning setup comprises 36 clients that each hold a subset of X-ray images from either the CheXpert or the Mendeley dataset. The clients represent hospitals or other medical institutions that provide their collected X-rays for the development of a classification model. The sizes of the clients' datasets are chosen such that they create a highly imbalanced setting including clients with very few data points, representing small institutions that make their limited amount of data available as soon as they are collected.

We simulated five clients with large subsets of the original CheXpert training set, and 31 clients with small subsets of either the original CheXpert validation set or the original Mendeley training set. We randomly split the patients whose images are part of the original CheXpert training set into five equal parts and assign each part randomly to one of five clients. Table \ref{table:dataset_splits} shows the distribution of the CheXpert validation data and Mendeley data among the remaining 31 clients, split in training, validation and test set sizes. These clients are used as targets for the reconstruction attacks in Sections~\ref{sec:results_attack} and \ref{sec:results_dp_attack}. Each client's dataset was further split into a dedicated training, validation, and test set, consisting of 70\%, 15\%, and 15\% of the client's data, respectively. Clients' datasets that are smaller than 50 images were split equally among the subsets. Datasets comprising less than ten images were used solely for training, omitting local validation or testing. All splits were performed randomly. No specific label distribution was enforced. We ensured that there was no patient overlap between clients and between training, validation, and test splits within each client's dataset.

\begin{specialtable}[h]
\caption{Number of images from the Mendeley training dataset (\subref{table:mendeley_split}) and Chexpert validation dataset (\subref{table:chexpert_split}), distributed among 14 and 17 clients, respectively. We specify how many clients are included that hold the respective amount of data. The last row shows the total of previous rows, taking into account the number of clients. }
\label{table:dataset_splits}
\centering
\captionsetup{justification=centering}
\begin{subtable}[t]{.5\linewidth}
\centering
\caption{Mendeley clients.}
\label{table:mendeley_split}
\begin{tabular}{c c c c | c}
\toprule
\makecell{\textbf{No.}\\\textbf{Clients}} & \textbf{Train} & \textbf{Val.} & \textbf{Test} & \textbf{Total}\\
\midrule
2 & 350 & 75 & 75 & 500\\
2 & 140 & 30 & 30 & 200\\
2 & 70 & 15 & 15 & 100\\
2 & 10 & 10 & 10 & 30\\
2 & 4 & 3 & 3 & 10\\
2 & 2 & 0 & 0 & 2 \\
2 & 1 & 0 & 0 & 1\\
\midrule
14 & 1,686 & 1,154 & 266 & 266
\\
\bottomrule
\end{tabular}
\end{subtable}%
\begin{subtable}[t]{.5\linewidth}
\centering
\caption{CheXpert clients.}
\label{table:chexpert_split}
\begin{tabular}{c c c c | c}
\toprule
\makecell{\textbf{No.}\\\textbf{Clients}} & \textbf{Train} & \textbf{Val.} & \textbf{Test} & \textbf{Total}\\
\midrule
2 & 10 & 10 & 10 & 30\\
5 & 4 & 3 & 3 & 10\\
5 & 2& 0 & 0 & 2\\
5 & 1 & 0 & 0 & 1\\
\midrule
17 & 125 & 55 & 35 & 35\\
\bottomrule
\end{tabular}
\end{subtable}
\end{specialtable}

\subsubsection{Model Training}
\label{sec:training}

We compared a densely connected network (DenseNet) \cite{huang2017DenselyConnected} and a residual network (ResNet) \cite{he2016DeepResidual} because both architectures have proven especially successful for the task of X-ray image classification~\cite{baltruschat2019ComparisonDeep, ke2021CheXtransferPerformance}. We monitored the local models' performance during training on their client's validation set. The global, aggregated model was evaluated using the average performance over the clients' validation sets. The client's test sets were held back for unbiased, internal evaluation of the final global model after training has finished. As the performance metric, we used the area under the receiver operating characteristics curve (AUC).


Both DenseNet121 and ResNet50 models were initialized with parameters pre-trained on ImageNet data. A fully connected layer with sigmoid activation and the adjusted number of output neurons replaced the original final classification layer. We modified the model architectures to accept one channel grayscale instead of three-channel RGB inputs to reduce unnecessary model complexity. To still leverage pre-trained model parameters, we summed the three-channel parameters of the first model layer to obtain new weights for the one-channel input. Images were resized to 224\texttimes224 pixels and normalized with ImageNet parameters adapted to grayscale color encoding by averaging over the input channels, yielding normalization parameters $\mu=0.449$ and $\sigma=0.226$. We did not apply any data augmentation methods.

Because multiple training rounds on the same dataset increase the risk of privacy leakage, we did not perform hyperparameter tuning and settled on standard hyperparameters. The training ran for at most 20 communication rounds. Each client participated in every round. We set the local batch size to ten and adapted it accordingly for clients with fewer than ten data points. To avoid overfitting, clients performed a single local epoch~\cite{kaissis2020SecurePrivacypreserving, sheller2019MultiinstitutionalDeep}. Early stopping was applied if the AUC value of the global model did not improve for five consecutive rounds. We minimized the binary cross-entropy loss using SGD with an initial learning rate of $1\mathrm{e}\text{-}2$. The learning rate was reduced by a factor of $0.1$ when reaching a performance plateau, i.e., after the AUC of the global model has not improved for three consecutive rounds. The global model with the highest mean AUC across all clients was selected as the best final model.

In private training, the privacy loss is difficult to track for some layer types. This includes active batch normalization layers, which are part of both DenseNet and ResNet architectures, as they create arbitrary dependencies between samples within a single batch  \cite{ioffe2015BatchNormalization}. We experimented with different model layer freezing techniques to avoid training batch normalization layers resulting in intractable privacy loss. We refer to rendering model layers untrainable as \textit{layer freezing}. We considered full model training (no layer freezing), freezing batch normalization layers, and freezing all layers but the final classification layer.

\subsection{Reconstruction Attack}
\label{sec:attack}

Federated learning enables model training on distributed data without the need for direct data sharing. However, while federated learning satisfies the principle of data minimization by eliminating the need for data transfer, it is not by itself sufficiently privacy-preserving. Sensitive information about the training data can be inferred from shared models, which has been demonstrated in a variety of privacy attacks including inference of class representatives~\cite{hitaj2017DeepModels}, property inference~\cite{melis2019ExploitingUnintended}, membership inference~\cite{shokri2017MembershipInference}, and sample reconstruction~\cite{zhu2019DeepLeakage}.

We assume the server to be an honest-but-curious adversary with full knowledge of the federated as well as the local training procedures~\cite{kairouz2021AdvancesOpen}. It correctly orchestrates and executes the required computations. However, it has white-box access to shared model parameters and can passively investigate them without interfering with the training process.

Reconstruction attacks aim at recovering data samples from trained model parameters. A disclosure implies a serious privacy risk as X-ray images may reveal information about the patient's identity~\cite{packhauser2021MedicalChest} and sensitive properties such as patient age~\cite{sabottke2020EstimationAge}. Since reconstruction attacks can be conducted with little auxiliary information and in a passive manner, it is a relevant vulnerability within our threat model.

The \textit{Deep Leakage from Gradients} (DLG) attack enables pixel-wise reconstruction of training images from the model gradients obtained during SGD~\cite{zhu2019DeepLeakage}. The attack comprises the following steps:

 \begin{enumerate}
 \item Randomly initialize some dummy input data $x'$ and dummy label $y'$.
 \item Fit the given initial model with the dummy data and obtain dummy gradients $\nabla \theta'$.
 \item {Quantify the difference between the original and the dummy gradient by using the Euclidean ($\ell_2$) distance as the cost function:

 \begin{equation} \label{eq:g_diff}
 d_{grad}=\| \nabla \theta'-\nabla \theta\|^2
 \end{equation}
 }

 \item Iteratively minimize the distance between the dummy and original gradients by adjusting the dummy input and label using the following objective:

 \begin{equation}\label{g_obj}
 x'^*, y'^*= \argmin_{x',y'}\| \nabla \theta'-\nabla \theta\|^2
 \end{equation}

 \item End the optimization process when the loss is sufficiently small, indicating complete reconstruction of the input data, or when reaching a maximum number of iterations.

\end{enumerate}

Following subsequent work, we used an improved version of the attack. We assume that labels can be reconstructed analytically \cite{zhao2020IDLGImproved} and restrict the optimization to the image data. We used a loss function based on the cosine similarity between original and dummy gradients and the Adam optimizer as proposed by \citet{geiping2020InvertingGradients}. The cosine similarity loss is defined as follows:

 \begin{equation} \label{eq:cosine_loss}
 d_{grad}=1-\frac{\langle \nabla \theta', \nabla \theta \rangle}{\| \nabla \theta'-\nabla \theta\|} + \alpha TV(\mathbf{x'}).
 \end{equation}

 \noindent $TV(\mathbf{x'})$ is the total variation of the dummy image $\mathbf{x'}$, with factor $\alpha$ as a small prior. The loss is minimized based on the sign of its gradient.

To evaluate the vulnerability of the local models to the DLG attack in our federated learning setting, we simulated an adversarial server that applies the reconstruction attack to model updates received from individual clients. We chose an arbitrary client holding one image from the Mendeley dataset to evaluate the impact of model layer freezing and attack time on image reconstruction quality. We then attacked other clients holding up to ten training images to demonstrate that they are also susceptible to a privacy breach. We conducted three trials per attack, initializing dummy images from a random normal distribution. We determined the best result as the trial with the lowest cosine similarity loss. The initial learning rate of the Adam optimizer for the attack was $1\mathrm{e}\text{-}1$. We adopted the strategy from \citet{geiping2020InvertingGradients} and reduced the learning rate by a factor of $0.1$ after $3/8$\textsuperscript{th}, $5/8$\textsuperscript{th}, and $7/8$\textsuperscript{th} of the maximum number of iterations. Each trial ran for 20,000 optimization steps. The total variation factor $\alpha$ for the cosine similarity loss was $1\mathrm{e}\text{-}2$. We inferred the model gradients by computing the absolute difference between original model parameters and local model parameters after local training.

For quantitative evaluation of attack success, we used the peak signal-to-noise ratio (PSNR), measured in the unit of decibels (dB):

\begin{equation}
PSNR=20\cdot\log_{10} \left(\frac{MAX_I}{\sqrt{MSE}}\right),
\end{equation}

where $MAX_I$ is the difference between the minimum and the maximum possible pixel value and $MSE$ is the mean squared error (MSE) between two images.

In addition to quantifying attack success with the PSNR measure, we demonstrate to what degree sensitive patient information can be derived from reconstructed X-ray images. Even if an individual cannot always be identified directly from a particular image, statistical knowledge about demographic information and other sensitive properties in a given dataset may lead to unwanted conclusions about individuals. We compare the performance of auxiliary models that predict demographic patient information from original and reconstructed X-rays. Because there is no demographic patient information available for the Mendeley data, we focus our evaluation on clients holding parts of the CheXpert dataset. We centrally trained two auxiliary ResNet50 models with the original CheXpert training data to predict patient sex and age. Sex was encoded as a binary category. The corresponding loss function for model training was binary cross-entropy. The loss function for age prediction was the MSE in years between the true and the predicted age. The sigmoid activation function in the age prediction model was replaced by a rectified linear unit (ReLU). Validation was carried out on a dedicated part of the CheXpert training data. The models achieved a validation AUC of $0.97$ on sex classification and a mean absolute error (MAE) of $6.0$ on age prediction. We applied the auxiliary classification on reconstructed images from clients that hold subsets of the original CheXpert validation data, thus ensuring that the model was only used for inference on images it has not been trained with.

\subsection{Differential Privacy}
\label{sec:background_dp}

\citet{dwork2006DifferentialPrivacy} originally proposed the notion of differential privacy in the context of database systems. Differential privacy guarantees that the amount of information revealed about any individual record during a query remains unchanged regardless of whether the record is included in the database at the time of the query or not. Put differently, the probability of receiving a specific output from a query on a database should be almost the same when an individual record is part of the database or not. \textit{Almost} means that the probabilities do not differ by more than a specific factor, which is captured by the \textit{privacy budget} or \textit{privacy loss} $\varepsilon$. In the context of machine learning, we regard model training as a function of a dataset equivalent to a query that runs on a database. Intuitively, differential privacy applied to machine learning means that training a model on a dataset should likely result in the same model that would be obtained when removing a single sample from the dataset.

The formal definition of ($\varepsilon, \delta$)-differential privacy is as follows:

\begin{equation}
Pr[\mathcal{M}(x) \in \mathcal{S}] \leq \exp({\varepsilon}) \cdot Pr[\mathcal{M}(y) \in \mathcal{S}] + \delta,
\end{equation}

where $\mathcal{M}(x)$ is the (randomized) query or function, $x$ and $y$ are parallel databases that differ in at most one entry, $\mathcal{S} \subseteq Range(\mathcal{M})$, and $\delta$ is a small term relaxing the guarantee, usually interpreted as the probability that it fails. A randomized mechanism $\mathcal{M}(x)$ can be obtained by adding noise to the original function drawn from a statistical random distribution, e.g., the Laplacian or the Gaussian distribution. The amount of noise necessary to achieve ($\varepsilon, \delta$)-differential privacy is scaled to the \textit{$\ell_2$-sensitivity} of the function, which is the maximum distance between the outputs of a function run on two parallel databases. Differential privacy has two important qualities important to its application to machine learning. The output of a differentially private random mechanism remains differentially private during the application of another data-independent function (closure under post-processing). The privacy loss can be analyzed cumulatively over several applications of a mechanism on the same database (composability). We use the variant of Rényi differential privacy, based on the Rényi divergence, in combination with a Gaussian noise mechanism that allows for a tighter estimate of the privacy loss over composite mechanisms than ($\varepsilon, \delta$)-differential privacy  \cite{mironov2017RenyiDifferential}.

Differentially-private stochastic gradient descent (DP-SGD) is commonly deployed for integrating differential privacy into model training \cite{abadi2016DeepLearning}. DP-SGD adds two main steps to the SGD algorithm:

\begin{enumerate}
\item Bounding the function's sensitivity by clipping per-sample gradient $\ell_2$-norms to a clipping value $C$.
\item Adding Gaussian noise to the gradient, scaled to the sensitivity enforced by Step 1.
\end{enumerate}

We applied DP-SGD locally during training at the clients' sites. Private training was limited to at most ten communication rounds. A privacy accountant tracked the $\varepsilon$-guarantees for a specified list of orders $\alpha$ of the Rényi divergence over communication rounds. This yields the optimal $(\alpha, \varepsilon)$-pair at the end, where $\varepsilon$ is the lowest bound on the privacy loss in combination with the respective $\alpha$. Because the differentially private mechanism is closed under post-processing, aggregation of private model parameters yields a private global model that does not incur a larger privacy loss on individual clients' data than upper bounded by local DP-SGD.

To investigate the relationship between privacy and model performance for our use case, we limited the privacy budget to $\varepsilon \in \{1,3,6,10\}$. We tracked $\alpha$ values in $[1.1, 10.9]$ in steps of $0.1$, and values in $[12,63]$ in steps of $1$.

If $\delta$ is equal to or greater than the inverse of the size of the dataset, it would allow for leakage of a whole record or data sample without violation of the privacy constraint~\cite{dwork2014AlgorithmicFoundations}. As this is unacceptable, $\delta$ should be smaller than the inverse of the dataset size \cite{kaissis2021EndtoendPrivacy}. Because the size of individual client's dataset varies, we determined $\delta$ as follows:

\begin{equation}\label{eq:delta_client}
\delta_k = \mathrm{min}(\frac{1}{\|x_k\|_1} \cdot 0.9, 10^{-2}),
\end{equation}

\noindent where $\|x_k\|_1$ is the number of data samples in the training dataset of client $k$. Because some clients' datasets are very small, which would lead to high probabilities for the privacy guarantee to be violated, we defined a minimum value of $\delta=10^{-2}$.

We bounded the sensitivity of the training function by clipping per-sample gradients. An effective bound is a compromise between excessive clipping, which leads to biased aggregated gradient estimates that do not adequately represent the underlying true gradient values, and a loose clipping bound that forces to add an exaggerated amount of noise to the gradients. We employed global norm clipping, i.e., gradients were clipped uniformly over the course of training. Abadi et al. propose to use the median of unclipped gradient $\ell_2$ norms~\cite{abadi2016DeepLearning}. We could not obtain unclipped gradient norms directly because the clients' datasets were not considered available for non-private training. As a solution, we ran a few epochs of non-private, centralized training on an auxiliary chest X-ray dataset with the same training parameters as in our federated learning scenario \cite{kaissis2021EndtoendPrivacy}. We used the original Mendeley test set, which was not part of any of the clients' datasets. We randomly picked 5\% of the dataset as validation and test sets to validate the training procedure. Because centralized training on the Mendeley test set converged quickly, we tracked the medians over the first three epochs. We obtained median gradient norms of $0.42$ (DenseNet121) and $0.62$ (ResNet50) for models with frozen batch normalization layers, and $1.24$ (DenseNet121) and $0.72$ (ResNet50) for models with all layers frozen but the final layer.

\subsection{Implementation}

The implementation of all experiments is based on PyTorch\footnote{\url{https://pytorch.org/}. Last accessed December 3, 2021.} version 1.9. It is available under \url{https://github.com/linev8k/cxr-fl-privacy}. For differentially private model training, we used Opacus\footnote{\url{https://opacus.ai/}. Last accessed November 28, 2021.} version 0.14.0. Our privacy attacks follow the implementation published by Geiping et al.\footnote{\url{https://github.com/JonasGeiping/invertinggradients}. Last accessed December 3, 2021.} Our modified version is available under \url{https://github.com/linev8k/invert-gradients-cxr}.

\section{Results}
This section follows a similar structure as the previous one. We first show the results of the federated learning baseline. Then, we go over the evaluation of the reconstruction attack on the system, where we analyze different factors that impact attack success. Finally, we assess the impact of differential privacy, first on the federated learning effectiveness, and then on the reconstruction attack.
\subsection{Federated Learning Baseline}
\label{sec:fl_results}

We trained both models on the binary classification of the „No Finding“ label. The best global models achieved an AUC value of
$0.935$ (DenseNet121) and $0.938$ (ResNet50). The average AUC was larger on clients holding Mendeley data
($0.96$ for DenseNet121 and $0.95$ for ResNet50) compared to clients with CheXpert data
($0.85$ for DenseNet121 and $0.87$ for ResNet50). These results confirm the ability of deep learning models to reach a high classification performance on the Mendeley dataset \cite{banerjee2020MultidiseasesClassification, kaissis2021EndtoendPrivacy}, even when trained in a heterogeneous setting.

Given the implications of different layer freezing techniques for privacy, we compared the outcomes of full model training
(no layer freezing), freezing batch normalization layers, and freezing all layers but the final classification layer (Table~\ref{table:sc3_test_results}).

For both models, the performance after full model training and training with frozen batch normalization layers was similar with a maximum difference in AUC of $0.022$ on the test sets. We conclude that freezing batch normalization layers did not impede model training in our setting. In contrast, rendering all layers untrainable except for the final layer significantly decreased performance. This confirms outcomes from previous work where this transfer learning technique was found to be inferior to including more layers in training updates \cite{baltruschat2019ComparisonDeep}.

\begin{specialtable}[h]
\centering
\caption{Mean AUCs of the best global DenseNet121 and ResNet50 models, evaluated on the clients' test sets. \textit{Batch norm.} refers to freezing of batch normalization layers, \textit{All but last} to freezing all parameters except for the final classification layer. Training with frozen batch normalization layers delivered similar results to full model training.}
\label{table:sc3_test_results}

\begin{tabular}{l c c c}

\toprule
\multirow{2}{*}{\textbf{Model}} & \multicolumn{3}{c}{\textbf{AUC}} \\
\cmidrule{2-4}
& No freezing & Batch norm. & All but last\\
\midrule
DenseNet121 & $0.947$ & $0.935$ & $0.714$ \\

ResNet50 & $0.916$ & $0.938$ & $0.813$ \\

\bottomrule
\end{tabular}
\end{specialtable}

\subsection{Reconstruction Attack}
\label{sec:results_attack}

We attacked local models of arbitrary clients with varying layer freezing techniques, attack time points and batch sizes.

\subsubsection{Impact of Layer Freezing}

\begin{figure}[h]
	\centering
	\begin{subfigure}[c]{0.2\linewidth}
	\includegraphics[width=\linewidth]{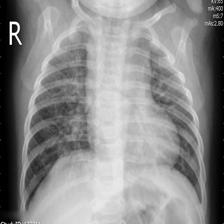}
	\caption{Original}
	\end{subfigure}
	\hspace{1em}
	\rotatebox{90}{ResNet}
	\begin{subfigure}[t]{0.2\linewidth}
	\includegraphics[width=\linewidth]{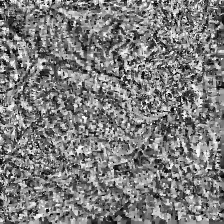}
	\end{subfigure}
	\hspace{1em}
	\begin{subfigure}[t]{0.2\linewidth}
	\includegraphics[width=\linewidth]{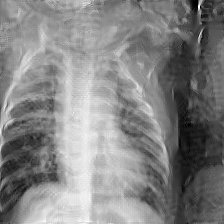}
	\end{subfigure}
	\hspace{1em}
	\begin{subfigure}[t]{0.2\linewidth}
	\includegraphics[width=\linewidth]{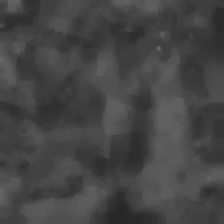}
	\end{subfigure}

	\vspace{-3\baselineskip}
	\hspace{0.2\linewidth}
	\hspace{1em}
	\rotatebox{90}{DenseNet}
	\begin{subfigure}[t]{0.2\linewidth}
	\includegraphics[width=\linewidth]{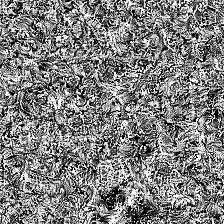}
	\caption{None}
	\label{fig:freezing_none}
	\end{subfigure}
	\hspace{1em}
	\begin{subfigure}[t]{0.2\linewidth}
	\includegraphics[width=\linewidth]{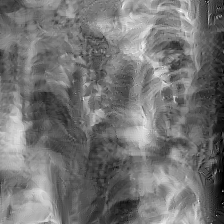}
	\caption{Batch norm.}
	\label{fig:bn_cl9}
	\end{subfigure}
	\hspace{1em}
	\begin{subfigure}[t]{0.2\linewidth}
	\includegraphics[width=\linewidth]{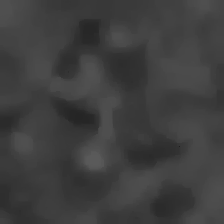}
	\caption{All but last}
	\label{fig:freezing_all_but_last}
	\end{subfigure}

	\caption{Best reconstructed images with varying layer freezing techniques. We attacked the locally trained model from a client holding a single Mendeley image. \textit{None} (\subref{fig:freezing_none}) refers to full model training, \textit{Batch norm.} (\subref{fig:bn_cl9}) to freezing batch normalization layers, and \textit{All but last} (\subref{fig:freezing_all_but_last}) to only training the output layer.}
	\label{fig:freezing_mend_attack}
\end{figure}

We applied the reconstruction attack to a single client's local model with a batch size of one during the first communication round. Table~\ref{table:fl_attack_results_freezing} reports the mean PSNR and sample standard deviation over three trials per experiment. Fig. \ref{fig:freezing_mend_attack} shows the reconstructed images of the best attack trials. The attack was only successful in the case of batch normalization layer freezing, indicated by larger mean PSNR values of $12.29$ (ResNet50) and $10.98$ (DenseNet121). Training the full model as well as fine-tuning only the output layer prevented the recovery of any useful image features in this setting. We further observed that the DenseNet121 seems to be more robust to leakage from gradients in this example, although the ResNet50 is the larger architecture in terms of parameter count, containing more than three times as many trainable parameters as the DenseNet121.

\begin{specialtable}[h]
\centering
\caption{Impact of layer freezing on the attack success during early training. We report the mean PSNR and sample standard deviation (STD) over all images obtained from three attack trials per setting. The batch size is kept constant at one. The attack was only successful on models with frozen batch normalization layers.}
\label{table:fl_attack_results_freezing}

\begin{tabular}{l c c c}
\toprule

\multirow{2}{*}{\textbf{Model}} & \multicolumn{3}{c}{\textbf{PSNR $\pm$ STD}} \\
\cmidrule{2-4}
 & None & Batch norm. & All but last \\
\midrule

ResNet50
& $9.73 \pm 0.09$ & $12.29 \pm 0.71$ & $8.50 \pm 0.12$ \\

DenseNet121 & $8.16 \pm 0.03$ & $10.98 \pm 0.18$ & $8.07 \pm 0.03$ \\

\bottomrule
\end{tabular}
\end{specialtable}

The results highlight that shared model updates with partial layer freezing are practically relevant targets for privacy violation. Cases of attack failure, however, do not provide a formal privacy guarantee. Other factors such as the privacy-breaking properties of active batch normalization layers in the case of full model training need to be considered for a comprehensive assessment of model privacy.

\subsubsection{Impact of Training Stage}

\begin{figure}[h]
	\centering
	\begin{subfigure}[c]{0.2\linewidth}
	\includegraphics[width=\linewidth]{images/mend_layer/cl9_gt.png}
	\caption{Original}
	\end{subfigure}
	\hspace{1em}
	\rotatebox{90}{ResNet}
	\begin{subfigure}[t]{0.2\linewidth}
	\includegraphics[width=\linewidth]{images/mend_layer/resnet_bn_cl9.png}
	\end{subfigure}
	\hspace{1em}
	\begin{subfigure}[t]{0.2\linewidth}
	\includegraphics[width=\linewidth]{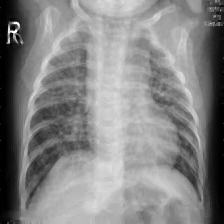}
	\end{subfigure}

	\vspace{-3\baselineskip}
	\hspace{0.2\linewidth}
	\hspace{1em}
	\rotatebox{90}{DenseNet}
	\begin{subfigure}[t]{0.2\linewidth}
	\includegraphics[width=\linewidth]{images/mend_layer/densenet_bn_cl9.png}
	\caption{Early training}
	\label{fig:bn_early}
	\end{subfigure}
	\hspace{1em}
	\begin{subfigure}[t]{0.2\linewidth}
	\includegraphics[width=\linewidth]{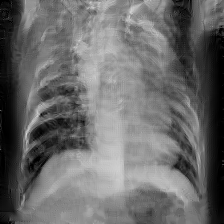}
	\caption{Late training}
	\label{fig:bn_late}
	\end{subfigure}

	\caption{Best reconstructed images after the first (\subref{fig:bn_early}) and after the fourth (\subref{fig:bn_late}) communication round. Reconstruction quality increased significantly after several rounds of training.}
	\label{fig:stage_mend_attack}
\end{figure}

We applied the attack during the initial communication round and after four rounds of training. We refer to the settings as an attack in \textit{early} and \textit{late} training stages, respectively. Fig. \ref{fig:stage_mend_attack} compares the images obtained from early and late attacks. The late attack was significantly more successful on both ResNet50 (mean PSNR $18.42 \pm 5.25$) and DenseNet121 (mean PSNR $11.7 \pm 2.23$). At the same time, the variation between trials was greater for the late attack.

\begin{figure}
  \centering
  \includegraphics[width=0.75\linewidth]{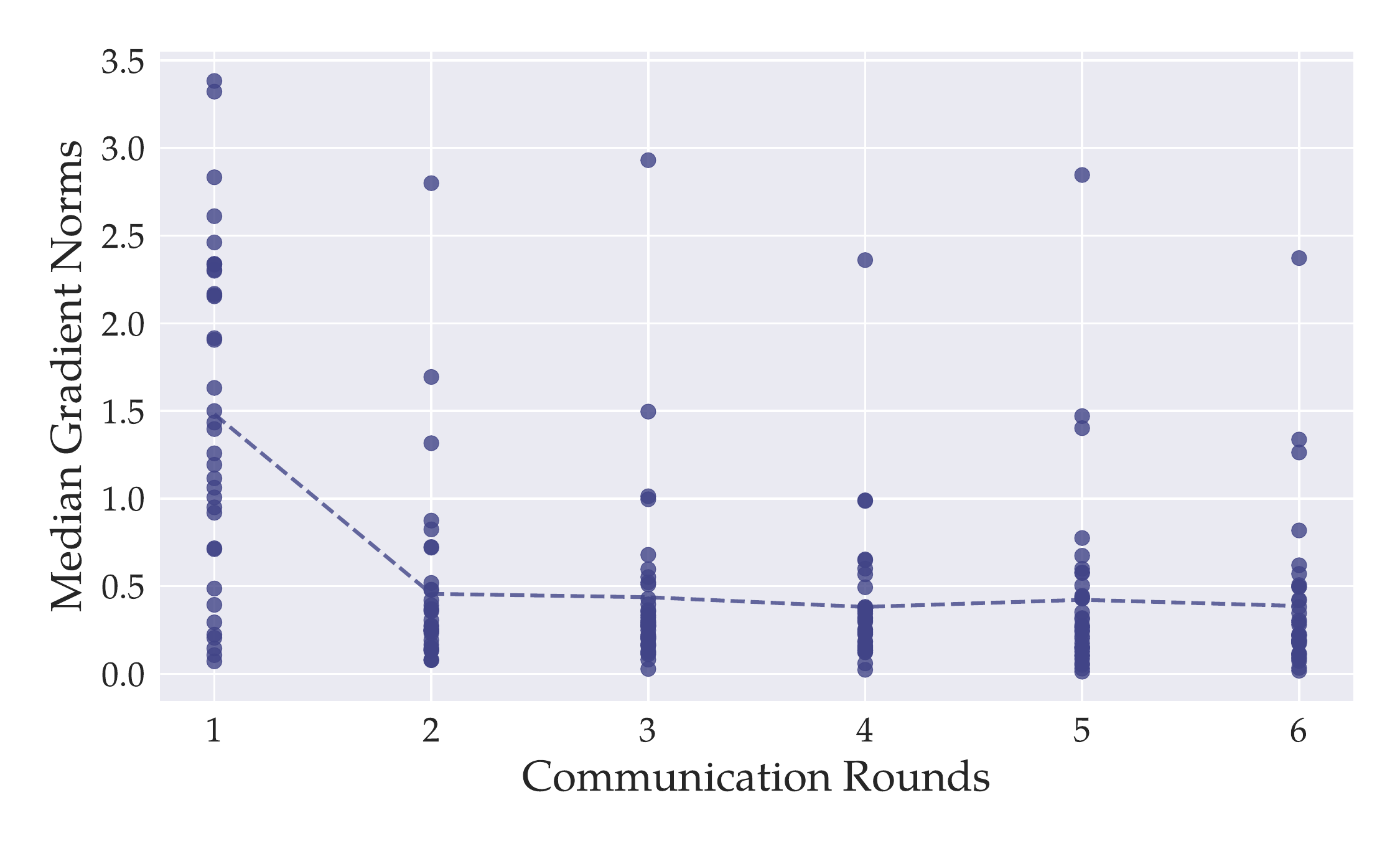}

  \caption{For each model layer of the DenseNet121, we tracked the median $\ell_2$-norms during training of every local model. Each dot represents the mean of one layer-median over all local models. The dashed line depicts the overall mean of all per-layer $\ell_2$-norm medians. Most layers' $\ell_2$-norms were greater during the first round of training than in later stages. In our experiments, image reconstruction was better on models from later rounds, suggesting that the magnitude of gradient $\ell_2$-norms is not a primary indicator for attack success.}
  \label{fig:l2norm_training}
\end{figure}

The observation that the attack was more successful as training progressed does not confirm previous evidence, which suggests that reconstruction is less successful from pre-trained models \cite{geiping2020InvertingGradients} and during later training stages \cite{kaissis2021EndtoendPrivacy}. Attack success has been associated with the magnitude of the gradients' $\ell_2$-norms, which are usually largest at the beginning when the model starts training on previously unseen data \cite{kaissis2021EndtoendPrivacy}. In Fig. \ref{fig:l2norm_training}, we investigate how the $\ell_2$-norms of our models' gradients changed as training progressed. We show the exemplary case of the pre-trained DenseNet121. Results were similar for the ResNet50, for which we refer to Appendix Section \ref{app:norm_resnet}. For each layer of every client's local model, we tracked the median $\ell_2$-norm during training. We display the per-layer mean values of all tracked medians over the local models. For both model architectures, the norms were greater during the first round of training than in the following iterations. Subsequent changes are more subtle and lack continuity. We validated that the attacked client's model did not pose an exception to this behavior.

Since our attacks were more successful during late training and we observed overall smaller gradient norms as training progressed, we could not associate larger gradient norms with increased attack success.

\subsubsection{Impact of Batch Size}

We investigated the impact of the training batch size in the setting where the attack was most successful, i.e., on models trained with frozen batch normalization layers attacked during late training. We attacked clients for which the considered batch size was equal to the available number of training images. The setting is equivalent to clients with larger datasets sharing model updates after every processed batch. A batch size of ten reduced attack success as the mean PSNR values over the batch decreased to $9.01$ (ResNet50) and $8.15$ (DenseNet121) (Table \ref{table:fl_attack_results_batch}). While the quality of the reconstructions varied for individual images within a batch, at least one image out of each batch became recognizable. We note that the order of images in a batch may not be preserved in the reconstruction of larger batches, preventing a direct comparison between original and reconstructed data points. To assign a reconstructed image to its original for evaluation, we first obtained the PSNR of each original image with each reconstructed image. We then determined the first original-reconstruction pair as the one with the largest PSNR value. The next best pair was determined considering the PSNR values between the remaining original and reconstructed images. We iterated the procedure until all images have been assigned.

\begin{figure}[h]
	\centering
	\rotatebox{90}{ResNet}
	\begin{subfigure}[t]{0.2\linewidth}
	\includegraphics[width=\linewidth]{images/mend_stage/resnet_bn_late.png}
	\end{subfigure}
	\hspace{1em}
	\begin{subfigure}[t]{0.2\linewidth}
	\includegraphics[width=\linewidth]{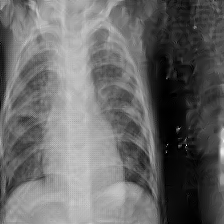}
	\end{subfigure}
	\hspace{1em}
	\begin{subfigure}[t]{0.2\linewidth}
	\includegraphics[width=\linewidth]{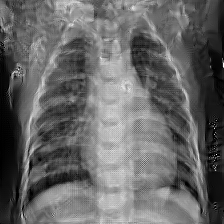}
	\end{subfigure}
	\hspace{1em}
	\begin{subfigure}[t]{0.2\linewidth}
	\includegraphics[width=\linewidth]{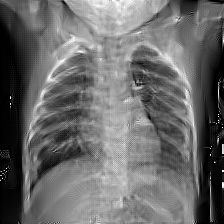}
	\end{subfigure}

	\vspace{0.2cm}
	\rotatebox{90}{DenseNet}
	\begin{subfigure}[t]{0.2\linewidth}
	\includegraphics[width=\linewidth]{images/mend_stage/densenet_bn_late.png}
	\caption{Batch size 1}
	\label{fig:bn_one}
	\end{subfigure}
	\hspace{1em}
	\begin{subfigure}[t]{0.2\linewidth}
	\includegraphics[width=\linewidth]{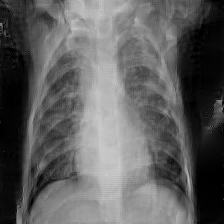}
	\caption{Batch size 2}
	\label{fig:bn_two}
	\end{subfigure}
	\hspace{1em}
	\begin{subfigure}[t]{0.2\linewidth}
	\includegraphics[width=\linewidth]{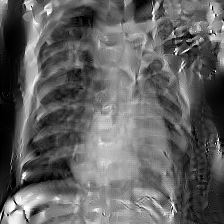}
	\caption{Batch size 4}
	\label{fig:bn_four}
	\end{subfigure}
	\hspace{1em}
	\begin{subfigure}[t]{0.2\linewidth}
	\includegraphics[width=\linewidth]{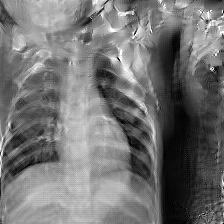}
	\caption{Batch size 10}
	\label{fig:bn_ten}
	\end{subfigure}

	\caption{Best reconstructed images out of each batch of the specified size. While other samples from those batches were not affected by the attack, the privacy of these examples' original X-rays has been severely breached, regardless of the batch size.}
	\label{fig:batch_mend_attack}
\end{figure}

Fig. \ref{fig:batch_mend_attack} shows the best-reconstructed images out of each batch, demonstrating that all considered batch sizes permit severe privacy breaches on individual data samples.

\begin{specialtable}[h]
\centering
\caption{Impact of batch size on attack success during late training. We report the mean PSNR and sample standard deviation (STD) over all images obtained from three attack trials per setting. Models were trained with frozen batch normalization layers (cf. Fig.~\ref{fig:freezing_mend_attack}). Attack success deteriorated with a batch size of ten, but not significantly with smaller batch sizes.}
\label{table:fl_attack_results_batch}
\begin{tabular}{l c c c c}
\toprule

\multirow{2}{*}{\textbf{Model}} & \multicolumn{4}{c}{\textbf{PSNR $\pm$ STD}} \\
\cmidrule{2-4}
& 1 & 2 & 4 & 10\\
\midrule

ResNet50
& $18.42 \pm 5.25$ & $11.79 \pm 1.2$ & $14.60 \pm 2.86$ & $9.01 \pm 3.13$\\

DenseNet121
& $11.7 \pm 2.23$
& $12.47 \pm 3.24$
& $12.67 \pm 2.24$
& $8.15 \pm 2.82$\\

\bottomrule
\end{tabular}
\end{specialtable}

\subsubsection{Inference of Demographic Properties}
\label{subsec:inference_properties}
Finally, to investigate the leakage of sensitive patient information from reconstructed images, we applied the attack to 15 clients holding CheXpert validation data subsets. We included five clients each, holding one, two, and four training images, yielding 35 images in total. The setting was the same as for the attacks on Mendeley clients. We attacked models trained with frozen batch normalization layers during late training. Then, we predicted the patients' age and sex from the original X-rays and from the reconstructed images using auxiliary models to demonstrate that the images leak sensitive information.

Table \ref{table:classif_chexpert} summarizes the auxiliary model predictions. The low baseline performance of the auxiliary models on original images compared to the classifier validation estimate is probably due to the small sample size of $35$ images. Superior results on images reconstructed from the ResNet50 in the case of sex prediction suggest an increased susceptibility to privacy violation of this architecture compared to the DenseNet121.

\begin{specialtable}[h]
\centering
\caption{Performance of the auxiliary models for predicting patient sex and age from X-ray images. We compare the classification/regression of original images, and images reconstructed from local ResNet50 and DenseNet121 models. All attacked clients provided $35$ images in total. Metrics reported are AUC for sex prediction and the mean absolute error (MAE) in years for age regression.}
\label{table:classif_chexpert}

\begin{tabular}{l c c}

\toprule
\multirow{2}{*}{\textbf{Attacked Model}} & \multirow{2}{*}{\shortstack{\textbf{Sex}\\(AUC)}} & \multirow{2}{*}{\shortstack{\textbf{Age}\\(MAE)}}\\ \\
\midrule

- &  $0.71$ & $11.51$\\
ResNet50 & $0.69$ & $15.42$\\
DenseNet121 & $0.56$ & $15.22$\\

 \bottomrule
\end{tabular}
\end{specialtable}

\subsection{Differentially Private Federated Learning}
\label{sec:results_dp}
As a countermeasure to the reconstruction attack, we evaluate the introduction of local differential privacy into our training process. The following sections detail the implications that come with that added protection.

\subsubsection{Model Performance}

Table \ref{table:privacy_model_results} reports the models' performance with privacy budgets $\varepsilon \in \{1,3,6,10\}$. We include the non-private baseline performance for comparison. Batch normalization layer parameters were not updated during model training. We report the exact privacy budget spent by each local model as optimal $(\alpha, \varepsilon)$-pairs in Appendix Section \ref{app:eps_alpha}.

\begin{specialtable}[h]
\centering
\caption{Mean AUC of global DenseNet121 and ResNet50 models, evaluated on the clients' test sets for non-private training and private training with varying $\varepsilon$ values. Stronger privacy guarantees decreased model performance.}
\label{table:privacy_model_results}

\begin{tabular}{l c c c c c}

\toprule
\multirow{2}{*}{\textbf{Model}} & \multicolumn{5}{c}{\textbf{AUC}} \\
\cmidrule{2-6}
& - & $\varepsilon=10$ & $\varepsilon=6$ & $\varepsilon=3$ & $\varepsilon=1$ \\
\midrule

DenseNet121 & $0.935$ & $0.925$ & $0.937$ & $0.854$ & $0.800$\\

ResNet50 & $0.938$ & $0.861$ & $0.764$ & $0.711$ & $0.622$\\

\bottomrule
\end{tabular}
\end{specialtable}

\begin{figure}[h]
	\centering
	\includegraphics[width=0.75\linewidth]{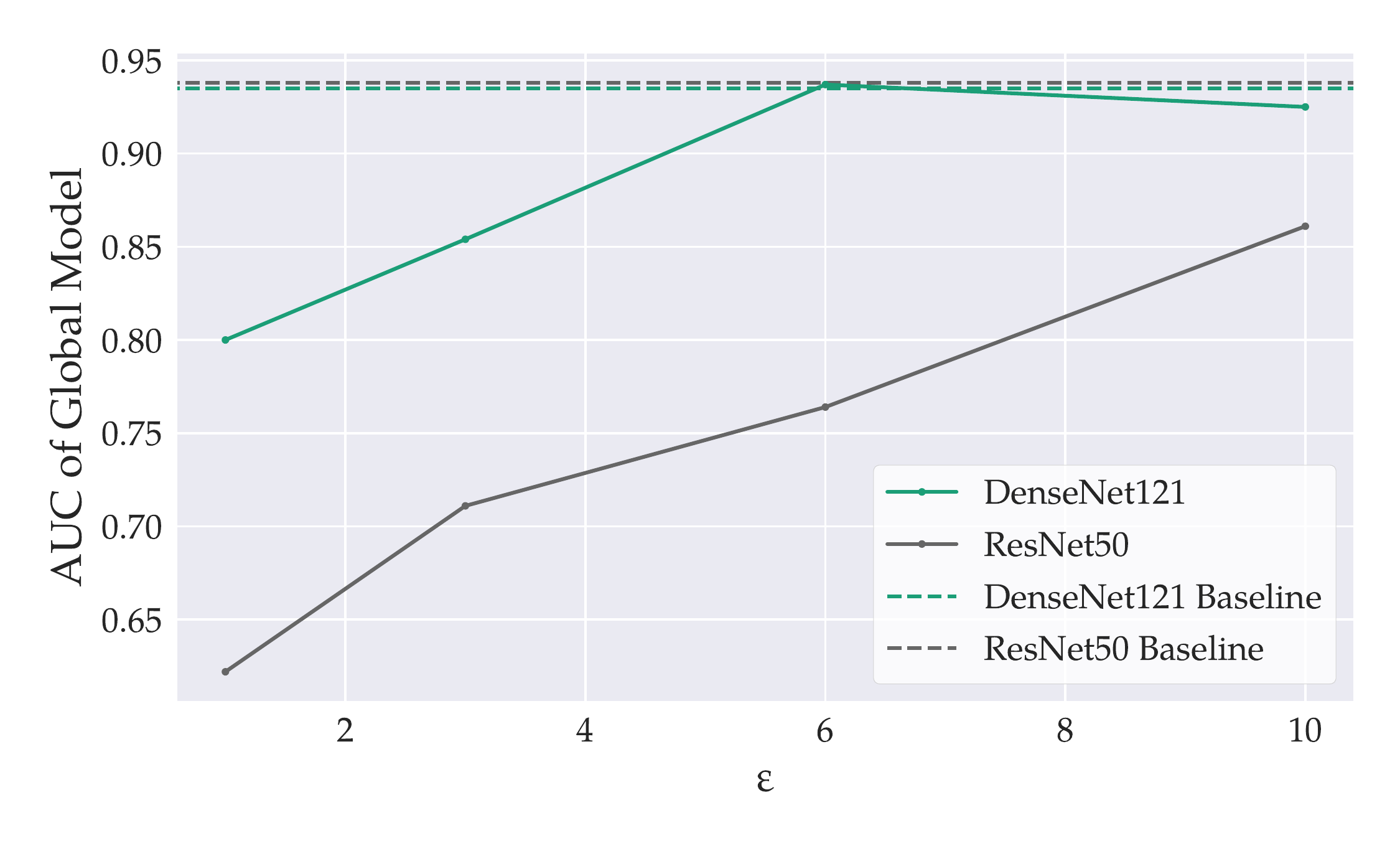}
	\caption{Model performance, evaluated on the clients' test sets in dependence on the privacy budget~$\varepsilon$. Baselines mark the peak performance of global non-private models. Stronger privacy guarantees degraded model accuracy.}
	\label{fig:dp_priv_perf}
\end{figure}

We compare the utility-privacy trade-off between the two model architectures in Fig.~\ref{fig:dp_priv_perf}. The DenseNet121 performed better than ResNet50 for all considered privacy budgets. As expected, a stronger privacy guarantee claimed a higher cost in accuracy for both models. The degradation was more pronounced in the ResNet50 with an AUC difference of $0.24$ between $\varepsilon=10$ and $\varepsilon=1$. The private DenseNet121 performed equally well compared to its non-private counterpart for both $\varepsilon=10$ and $\varepsilon=6$, suggesting that a increasing the privacy budget beyond $\varepsilon=6$ does not benefit model performance. For $\varepsilon=6$, the DenseNet121 achieved an AUC of $0.937$, the ResNet50 only $0.764$.

We expect that model evaluation in our setting with imbalanced data distribution tends to be unreliable on clients with less data. Incidental good results on those clients may bias the global model's performance estimate. To provide a more meaningful assessment of the model performance under privacy conditions, we investigated the performance of the best global private DenseNet121 model on individual clients compared to the best non-private model. We visualize the comparison for $\varepsilon=6$ in Fig.~\ref{fig:clients_comp_eps}. We provide the figures for other considered $\varepsilon$-values in Appendix Section \ref{app:clients_eps}. Private training demanded a systematic cost in performance for clients holding large amounts of CheXpert data. AUC values on those clients' datasets decreased by $0.03$ ($\varepsilon=10$ and $\varepsilon=6$) and $0.06$ ($\varepsilon=3$) on average from non-private to private training.

We conclude that the impact of private training on model accuracy, also at moderate privacy budgets, needs to be carefully assessed on the client level. Further potential weak points of the resulting model, such as performance on underrepresented patient subgroups, require additional consideration.

\begin{figure}[h]
  \centering
    	\includegraphics[width=0.75\linewidth]{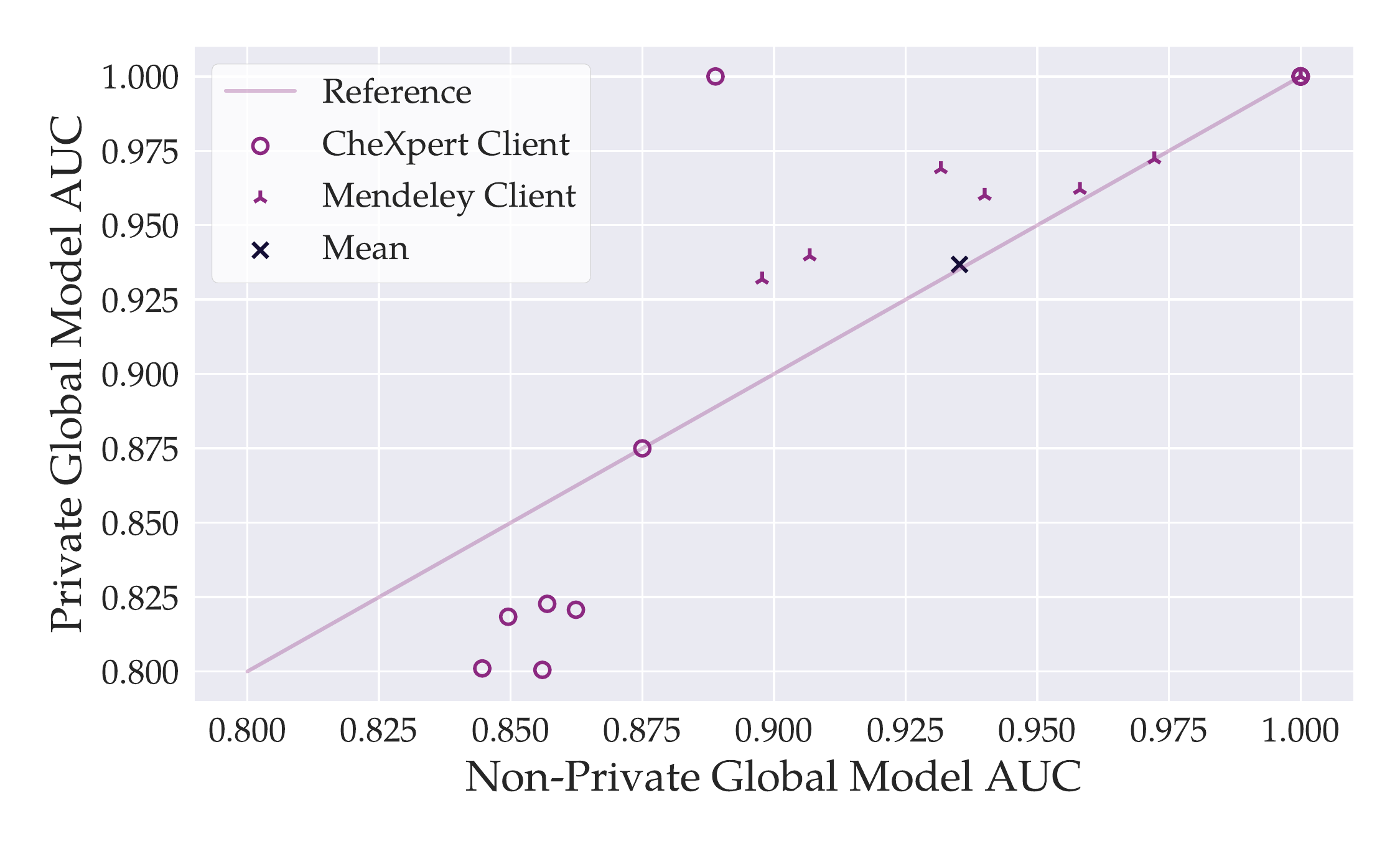}

  \caption{Comparison of per-client AUC values achieved by the best global DenseNet121 model between private ($\varepsilon=6$) and non-private training. The bottom left circle markers belong to CheXpert clients with large datasets. Privacy demanded a higher cost in model accuracy on CheXpert clients. The figure does not consider AUC values below $0.79$. Markers may overlap.}

 \label{fig:clients_comp_eps}
\end{figure}

\subsubsection{Additional Training Techniques}

We evaluate the effect of additional training techniques on model performance: Training only the final layer (\textit{All but last} layer freezing), client subsampling, and employment of layer-wise gradient clipping. All experiments were carried out with a privacy budget of $\varepsilon=10$. We found that none of the techniques introduced an advantage for private model training. When restricting training to the final layer, the performance of the DenseNet121 decreased significantly compared to \textit{Batch norm.} freezing (AUC $0.707$ vs. $0.925$) and that of the ResNet50 remained similar (AUC $0.871$ vs. $0.861$).

In a separate experiment, we introduced a client subsampling procedure where the maximum number of global communication rounds was set to ten and the maximum number of rounds that each client can be selected to five. The fraction of clients chosen each round was $0.3$, resulting in eleven clients selected per round. This way, less of the available privacy budget was effectively spent during the clients' local training because each client participated in fewer training rounds in total. However, there was no improvement in model performance. The DenseNet121 and ResNet50 achieved AUC values of $0.836$ and $0.819$, respectively. A potential explanation is that clients with small datasets got selected frequently during subsampling, but could not contribute as effectively to the global model as clients with larger datasets. Model accuracy degraded more heavily after a few rounds during the subsampling experiment, indicating stronger local overfitting which was amplified by the lower number of contributing clients.

Finally, instead of uniformly clipping the norm of each gradient value, we specified an individual clipping bound for each model layer. We utilized the per-layer median gradient norms from the auxiliary training experiment on the Mendeley test set (Section \ref{sec:background_dp}). The models' AUC values converged to $0.5$, indicating that model training failed for our use case when employing layer-wise gradient clipping. The variation between individual clipping values may be too large, preventing the model parameters to retain any information that is usable in combination with other layers' parameters.

\begin{figure}
	\centering
	\rotatebox{90}{ResNet}
	\begin{subfigure}[t]{0.2\linewidth}
	\includegraphics[width=\linewidth]{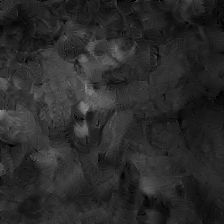}
	\end{subfigure}
	\hspace{1em}
	\begin{subfigure}[t]{0.2\linewidth}
	\includegraphics[width=\linewidth]{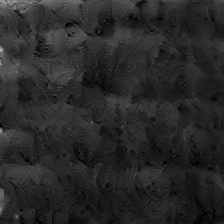}
	\end{subfigure}
	\hspace{1em}
	\begin{subfigure}[t]{0.2\linewidth}
	\includegraphics[width=\linewidth]{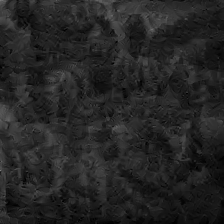}
	\end{subfigure}
	\hspace{1em}
	\begin{subfigure}[t]{0.2\linewidth}
	\includegraphics[width=\linewidth]{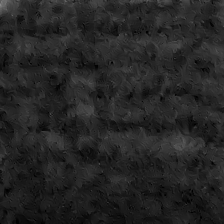}
	\end{subfigure}

	\vspace{0.2cm}
	\rotatebox{90}{DenseNet}
	\begin{subfigure}[t]{0.2\linewidth}
	\includegraphics[width=\linewidth]{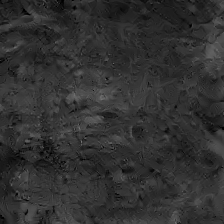}
	\caption{$\varepsilon=10$}
	\end{subfigure}
	\hspace{1em}
	\begin{subfigure}[t]{0.2\linewidth}
	\includegraphics[width=\linewidth]{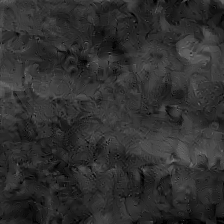}
	\caption{$\varepsilon=6$}
	\end{subfigure}
	\hspace{1em}
	\begin{subfigure}[t]{0.2\linewidth}
	\includegraphics[width=\linewidth]{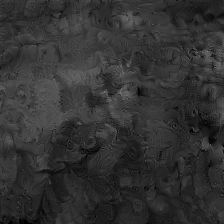}
	\caption{$\varepsilon=3$}
	\end{subfigure}
	\hspace{1em}
	\begin{subfigure}[t]{0.2\linewidth}
	\includegraphics[width=\linewidth]{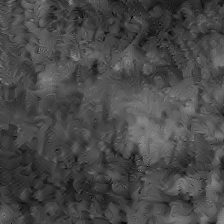}
	\caption{$\varepsilon=1$}
	\end{subfigure}

	\caption{Best reconstructed images under different privacy constraints. Private training successfully prevents the leakage of any visible features.}
	\label{fig:priv_rec_images}
\end{figure}

\subsubsection{Vulnerability to Reconstruction Attack}
\label{sec:results_dp_attack}
We attempted to reconstruct the training image from the local model shared by a Mendeley client during private training. We performed the attack on models with frozen batch normalization layers during late training. Table \ref{table:priv_attack_results} compares the mean PSNR over three trials between non-private and private training. The PSNR on all images from private models was significantly smaller than in the non-private setting. Fig. \ref{fig:priv_rec_images} confirms that the reconstructed images from both model architectures did not leak any visual parts of the training images. Differentially private training under all considered privacy budgets therefore successfully prevented the attack.

\begin{specialtable}[h]
\centering
\caption{Mean PSNR over three attack trials on non-private and private local models from a Mendeley client holding one training image. The attack failed for all considered $\varepsilon$ values.}
\label{table:priv_attack_results}

\begin{tabular}{l c c c c c}

\toprule
\multirow{2}{*}{\textbf{Model}} & \multicolumn{5}{c}{\textbf{PSNR $\pm$ STD}} \\
\cmidrule{2-6}
& - & $\varepsilon=10$ & $\varepsilon=6$ & $\varepsilon=3$ & $\varepsilon=1$ \\
\midrule

DenseNet121 & $10.98 \pm 0.18$
& $6.58 \pm 0.07$
& $6.50 \pm 0.08$
& $6.46 \pm 0.07$
& $6.41 \pm 0.03$ \\

ResNet50 & $12.29\pm 0.71$
& $7.22 \pm 0.04$
& $7.13 \pm 0.16$
& $7.12 \pm 0.17$
& $8.49 \pm 0.32$ \\

\bottomrule
\end{tabular}
\end{specialtable}

To validate that no sensitive information was leaked, we applied the auxiliary models (first introduced in Section~\ref{subsec:inference_properties}) to predict patient age and sex from images reconstructed from private models. Table \ref{table:classif_chexpert_priv} compares their performance on original and recovered images in private and non-private settings. We attacked the model with the weakest privacy guarantee of $\varepsilon=10$. The AUC values of $0.49$ and $0.47$ on sex prediction indicate that the classifier's performance was equivalent to random label assignment in the private setting. The age predictions deviated around $19$ years on average from the true patients' age. Differentially private model training prevented both auxiliary models to predict usable information about the patients' demographic properties.

\begin{specialtable}[h]
\centering

\caption{Performance of the auxiliary models for predicting patient sex and age from X-ray images. We compare the predictions on original images, and images reconstructed from local ResNet50 and DenseNet121 models in the non-private and private setting with $\varepsilon=10$. Images reconstructed from private models leaked no usable information about the selected properties.}
\label{table:classif_chexpert_priv}

\begin{tabular}{l c c c}

\toprule
\multirow{2}{*}{\textbf{Attacked Model}} &
\multirow{2}{*}{$\varepsilon$} &
\multirow{2}{*}{\shortstack{\textbf{Sex}\\(AUC)}} & \multirow{2}{*}{\shortstack{\textbf{Age}\\(MAE)}}\\\\
\midrule

- &  - & $0.71$ & $11.51$ \\
\multirow{2}{*}{ResNet50} & - & $0.69$ & $15.42$ \\
& $10$ & $0.49$ & $19.23$\\
\multirow{2}{*}{DenseNet121} & - & $0.56$ & $15.22$\\
& $10$ & $0.47$ & $18.82$\\

\bottomrule
\end{tabular}
\end{specialtable}

We conclude, that in our federated learning setting, differential privacy is an effective countermeasure against sample reconstruction from gradients, and no sensitive information could be inferred from the reconstructed images.

\section{Discussion}
\label{sec:discussion}

In our federated learning setup, the effectiveness of model aggregation is limited by data heterogeneity and imbalance. The federated averaging algorithm weights the local model updates with respect to the clients' dataset size in relation to the overall amount of available data \cite{mcmahan2017CommunicationEfficientLearning}. This led to a strong emphasis on model updates from clients with large CheXpert subsets in our case, while updates from clients with fewer images contributed less to model aggregation. One option to mitigate data imbalance is to aggregate models after a specified number of batches instead of local epochs \cite{kaissis2021EndtoendPrivacy, wang2019AdaptiveFederated}. However, sharing intermediate models more frequently will increase the susceptibility to reconstruction attacks since the updates are obtained on small batches rather than the client's whole dataset. Improving the aggregation process under consideration of privacy costs is left for future work.

The applied attack has shown that the two considered deep machine learning models are susceptible to reconstruction of sensitive data from gradients. Most notably, and contrary to previous work, we found that the attack was more successful in later training stages and for pre-trained models.
Our privacy evaluation framework is limited by the choice of the \textit{Deep Leakage from Gradients} attack as a qualitative measure for model vulnerability. Even though we found reconstruction not successful under certain conditions, including full model training, restricting training to the final layer, and attacking the DenseNet121 at an early training stage, it cannot be assumed that model training would be privacy-preserving in these cases. Minor modifications of the attack scheme may improve attack success even in supposedly safe settings. Moreover, reconstruction attacks are only one example among a range of deliberate privacy breaches that neural networks are vulnerable to. Extending our privacy evaluation framework to include other privacy threats, e.g., property inference without data reconstruction, will provide further insights into potential vulnerabilities of the federated learning paradigm. Since the main limitation of \textit{Deep Leakage from Gradients} is its restriction to small datasets, it will be particularly valuable to capture the consequences of privacy breaches for clients with large amounts of data. From a security perspective, demonstrating that these attacks are practically feasible, albeit under limited circumstances, is sufficient for considering the machine learning process vulnerable to privacy violation. Countermeasures must constantly be re-evaluated for their effectiveness as a better understanding of privacy threats evolves.

Our privacy evaluation was further constrained to a limited choice of privacy budgets. While choosing $\varepsilon=6$ delivered the best utility-privacy trade-off for our use case, which is in line with previous work \cite{kaissis2021EndtoendPrivacy}, it may not be the optimal lower bound. We specifically suggest empirically examining choices of $\varepsilon$ in the range $[3,6]$ to potentially improve upon our results in future work. We also note that while all considered privacy constraints prevented the success of reconstruction attacks, smaller $\varepsilon$ values still formally provide stronger privacy guarantees that offer protection against threats beyond the limited case of the reconstruction attack.

A key implication of our results is that the DenseNet121 architecture proved more robust against private model training with regard to performance than the ResNet50. This observation is potentially related to the greater ability of the DenseNet121 to withstand reconstruction attacks. Although the model contains overall fewer parameters than the ResNet50, its dense structure may, to a certain degree, offer a natural defense against reconstruction from trained parameters as well as perturbation of parameter updates during private training. This outcome suggests substantial differences in the suitability of individual model types for privacy-preserving machine learning, which requires further validation.

In the medical context, fairness is crucial for the safe deployment of machine learning algorithms. Rare diseases or conditions must be reliably detected despite the restricted availability of representative data. Furthermore, a model should perform with equal accuracy on all patient subgroups. We uncovered performance differences between individual clients' data in our federated learning baseline, revealing that the classification produced better results on Mendeley than on CheXpert data. This potentially reflects the ability of deep learning models to recognize pneumonia as an abnormal finding particularly well since the pathologic X-rays in the Mendeley dataset only include cases of pneumonia. More thorough investigations are required to reveal other potential biases, e.g., with respect to patient subgroups. It is further known that underrepresented classes and population subgroups are potentially affected more strongly by model performance degradation when applying differential privacy~\cite{bagdasaryan2019DifferentialPrivacy}. Because model performance evaluation on Mendeley clients was less reliable due to smaller amounts of data, it remains an open question how exactly these clients were affected by the integration of differential privacy. For practical applications, it is mandatory to thoroughly investigate how privacy mechanisms affect the model's performance on different types of data to identify a potential underlying bias.

We did not consider the practical implementation of federated learning between different institutions with regard to communication time, required infrastructure, costs and validation of correct computational execution. The focus of our paper lies in analyzing the threat of data reconstruction and the effectiveness of differential privacy against it. While simulated use cases like ours are vital to prepare for leveraging differential privacy in real-world cases where sensitive data is involved, further case studies are required to investigate aspects of practicability for privacy-preserving federated learning on a large scale.

\section{Conclusions and Future Work}
\label{sec:conclusion}

We simulated a collaborative machine learning use case in which 36 institutions provide their diverse chest X-ray data collections for the development of a classification model. Two main concerns in this scenario are the physical separation of the data sources and the privacy of patients to whom the data belongs. We employed the paradigm of federated learning as a solution for machine learning on dispersed data. Throughout our experiments, we compared two large network architectures: DenseNet121 and ResNet50. Extending previous evidence, we demonstrated that individual X-rays can be reconstructed from shared model updates within the federated learning setting from those networks using the \textit{Deep Leakage from Gradients}. The attack is especially successful during later training stages.

As a step towards privacy-preserving distributed learning, we integrated Rényi differential privacy with a Gaussian noise mechanism into the federated learning process. The DenseNet121 achieved the best utility-privacy trade-off with a mean AUC of $0.937$ for $\varepsilon=6$, where we identified an expected cost in accuracy of $0.03$ in terms of the AUC on CheXpert clients' data compared to the non-private baseline. The results suggest that $\varepsilon \in [3,6]$ are suitable candidates for private model training depending on the specific demands on model privacy and performance for the respective application. Overall, we found the DenseNet121 model superior to ResNet50 with regard to private model training for all considered $\varepsilon$ values.

The adverse impact of differential privacy on model performance must be carefully considered, particularly for medical use cases. Our results endorse that differentially private federated learning is feasible at a small cost in model accuracy for the classification of heterogeneous chest X-ray data. As real-world medical use cases become more complex in practice, future work may elaborate on the potential of differentially private federated learning for multi-label X-ray classification where heterogeneous data from a broader range of sources is effectively integrated under consideration of an improved bound on the privacy budget. We identified the DenseNet121 as a robust model architecture suitable for differentially private training. Further comparison with other neural network architectures may reveal key indicators for the suitability of different model types and provide guidance in the choice of models for privacy-preserving machine learning. We further suggest to extend our evaluation framework in future work to consider the vulnerability to other types of privacy breaches, enabling a comprehensive qualitative assessment of model privacy. Finally, other variants of differential privacy, e.g. Gaussian differential privacy \cite{dong2019GaussianDifferential}, may offer suitable alternatives to the application of Rényi differential privacy providing yet tighter bounds on the privacy loss.




\vspace{6pt}



\clearpage
\authorcontributions{Conceptualization, J.Z., B.P., H.S., A.S. and B.A.; methodology, J.Z.; software, J.Z.; validation, J.Z. and B.P.; formal analysis, J.Z.; investigation, J.Z.; resources, J.Z., B.P. and B.A.; data curation, J.Z.; writing---original draft preparation, J.Z.; writing---review and editing, J.Z., B.P., H.S., A.S. and B.A.; visualization, J.Z.; supervision, B.P., H.S., A.S. and B.A.; project administration, B.A., B.P. and J.Z.; funding acquisition, B.A. \\All authors have read and agreed to the published version of the manuscript.}

\funding{This research received no external funding.}




\dataavailability{The CheXpert dataset \cite{irvin2019CheXpertLarge} is a public chest X-ray dataset available at \url{https://stanfordmlgroup.github.io/competitions/chexpert/}. The Mendeley dataset version 3 can be found at \url{https://data.mendeley.com/datasets/rscbjbr9sj/3}.}

\acknowledgments{Funded by the Deutsche Forschungsgemeinschaft (DFG, German Research Foundation) – Projektnummer 491466077}




\abbreviations{The following abbreviations are used in this manuscript:\\

\noindent
\begin{tabular}{@{}ll}
AUC & Area under the receiver operating characteristics curve\\
DLG & Deep Leakage from Gradients\\
DP-SGD & Differentially private stochastic gradient descent\\
MAE & Mean absolute error\\
MSE & Mean squared error\\
PSNR & Peak signal-to-noise ratio\\
ReLU & Rectified linear unit\\
ROC & Receiver operating characteristic\\
SGD & Stochastic gradient descent
\end{tabular}}

\appendixtitles{yes} 
\appendixstart
\appendix

\section{Gradient $\ell_2$-Norms}
\label{app:norm_resnet}

\begin{figure}[h]
  \centering
  \includegraphics[width=0.6\linewidth]{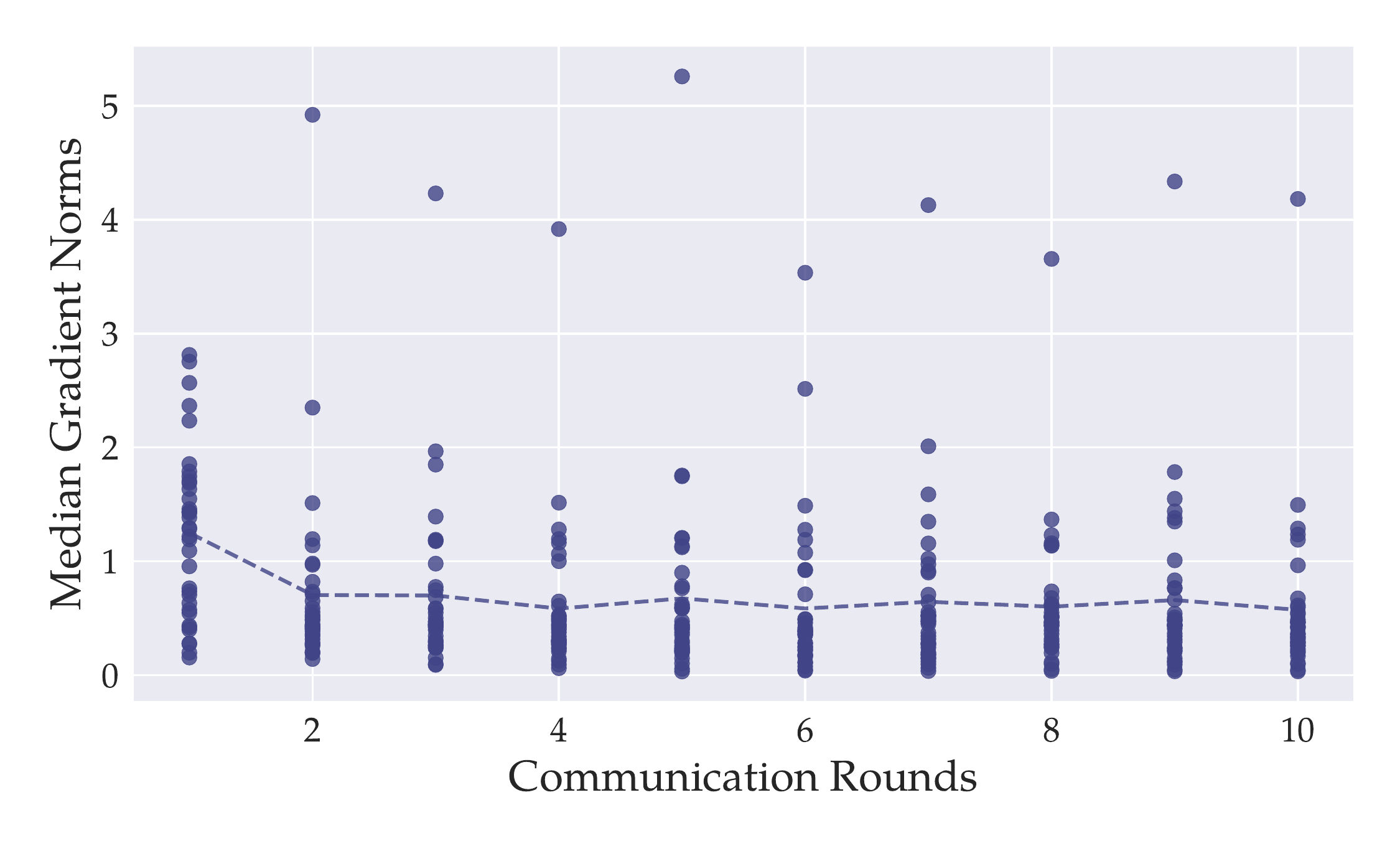}

  \caption{For each model layer in the ResNet50, we tracked the median $\ell_2$-norms during training of every local model. Each dot represents the mean of one layer-median over all local models. The dashed line depicts the overall mean of all per-layer $\ell_2$-norm medians. Most layers' $\ell_2$-norms were greater during the first round of training than in later stages.}
  \label{fig:l2norm_training_res}
\end{figure}

We investigated the models' gradients' $\ell_2$-norms during training to assess how they correlate with attack success. Fig. \ref{fig:l2norm_training_res} shows how the norms change in the ResNet50. The norms were greater during the first round of training than in the following iterations. In our experiments, image reconstruction was better on models from later rounds, suggesting that the magnitude of gradient $\ell_2$-norms is not a primary indicator for attack success.

\section{Client-level AUC of Private DenseNet121 Models}
\label{app:clients_eps}

\begin{figure}[h]
  \centering
  \captionsetup[subfigure]{justification=centering}
  \begin{subfigure}[c]{0.5\linewidth}
    \includegraphics[width=\linewidth]{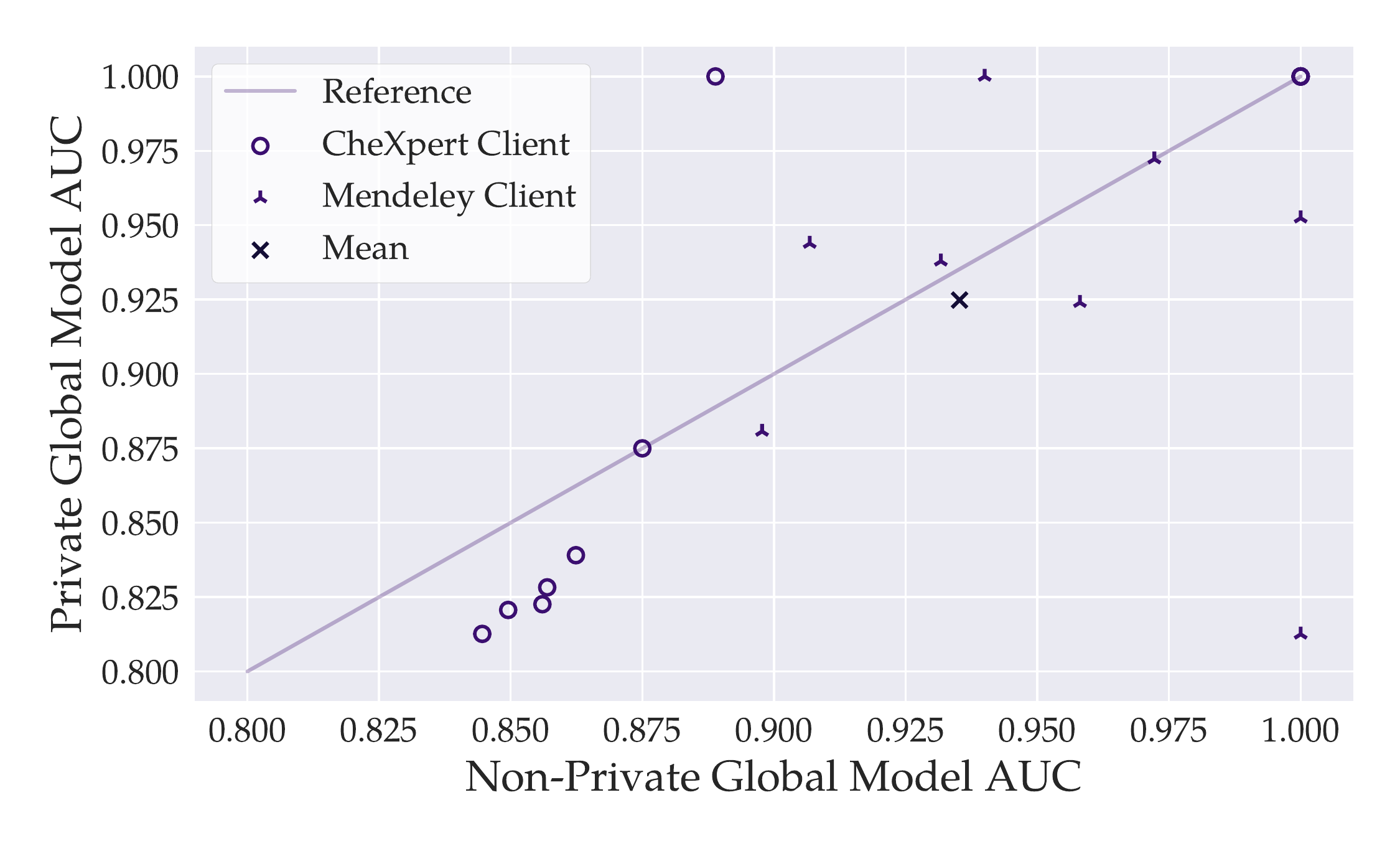}
    \caption{$\varepsilon=10$}
    \end{subfigure}

    \vspace{1em}

    \begin{subfigure}[c]{0.5\linewidth}
    	\includegraphics[width=\linewidth]{images/clients_comp_dense_eps6.pdf}
    \caption{$\varepsilon=6$}
    \end{subfigure}

    \vspace{1em}

    \begin{subfigure}[c]{0.5\linewidth}
    	\includegraphics[width=\linewidth]{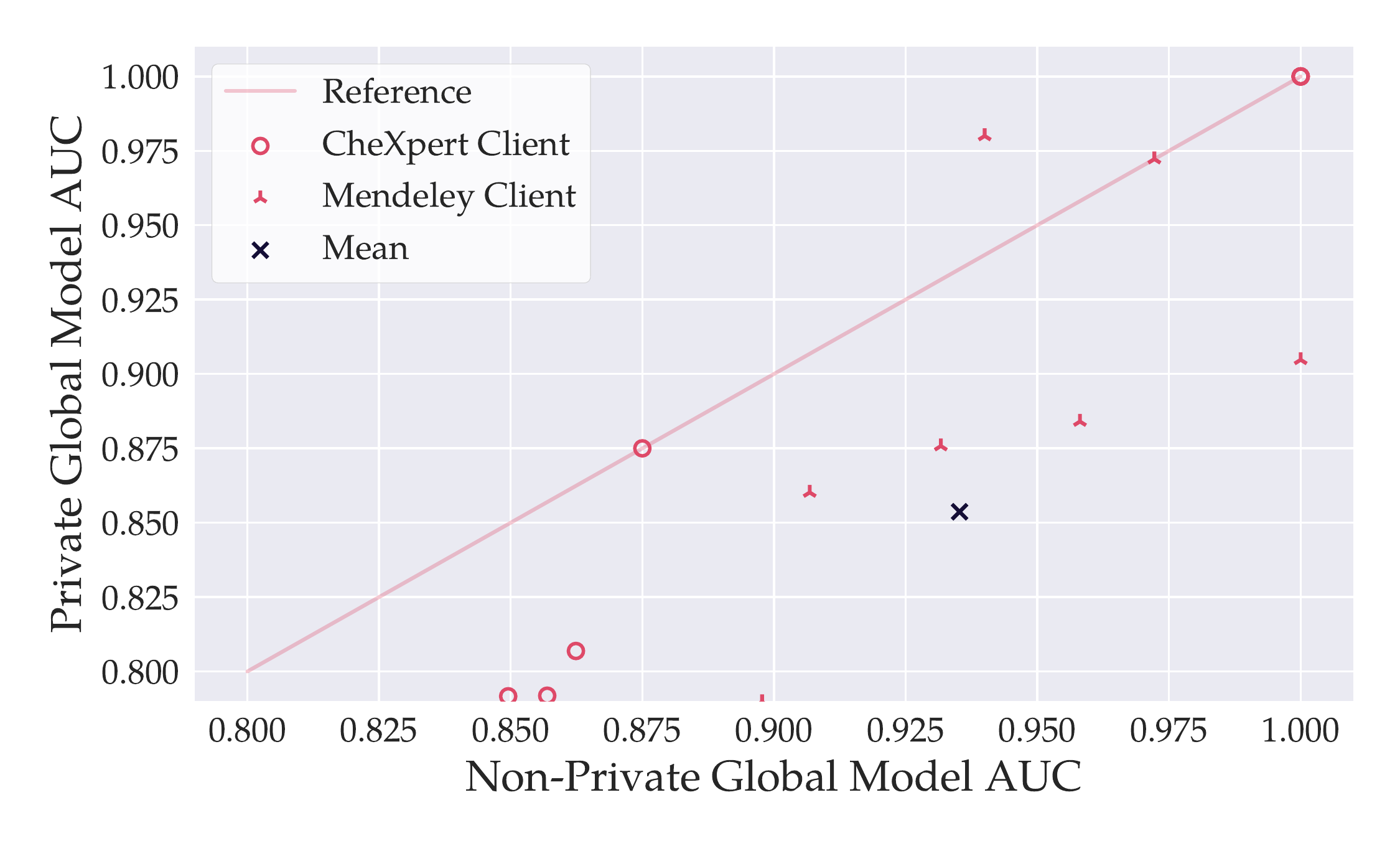}
    \caption{$\varepsilon=3$}
    \end{subfigure}

  \caption{Comparison of per-client AUC values achieved by the best global DenseNet121 models between private and non-private training. The bottom left circle markers belong to CheXpert clients with large datasets. Privacy demanded a higher cost in model accuracy on CheXpert clients. The figures do not consider AUC values below $0.79$. Markers may overlap.}

 \label{fig:clients_comp_eps_all}
\end{figure}

We investigated the performance of the best global private DenseNet121 model on individual clients compared to the best non-private model. We visualize the comparison for $\varepsilon \in {3,6,10}$ in Fig. \ref{fig:clients_comp_eps_all}. Because training was unsuccessful for $\varepsilon=1$, we do not evaluate model performance for this case in detail. Private training demanded a systematic cost in performance for clients holding large amounts of CheXpert data for all considered privacy budgets. Because clients with Mendeley data hold fewer images, the results on individual test sets of those clients was subject to greater variation.

\section{List of the Private Models' $(\alpha,\varepsilon)$-Pairs}
\label{app:eps_alpha}

Table \ref{table:eps_alpha} reports the exact privacy budget spent by each local model trained with differential privacy as optimal $(\alpha, \varepsilon)$-pairs.

\begin{specialtable}[h!]
\caption{Optimal $(\alpha, \varepsilon)$-pairs under fixed privacy budgets for each client's local model of our private training baseline. True $\varepsilon$ values may deviate slightly from the defined privacy budget as the noise multiplier is estimated  before model training in order to meet the privacy constraint. Clients 0 to 13 hold Mendeley data subsets, clients 14 to 18 the CheXpert training data, and clients 19 to 35 parts of the CheXpert validation data.}

\label{table:eps_alpha}

\centering
\begin{tabular}{c c c c c c}

\toprule

\multirow{2}{*}{\textbf{Client}} & \multirow{2}{*}{\shortstack{\textbf{No.}\\\textbf{Images}}} & \multicolumn{4}{c}{$(\alpha,\varepsilon)$} \\
\cmidrule{3-6}

& & $\varepsilon=10$ & $\varepsilon=6$ & $\varepsilon=3$ & $\varepsilon=1$ \\

\midrule

0 & 350 & $(1.9, 10.49)$ &$(2.4, 6.12)$ & $(3.7, 2.87)$&$(7.9, 1.01)$\\
1 & 350 & $(1.9, 10.49)$&$(2.4, 6.12)$ &$(3.7, 2.87)$ &$(7.9, 1.01)$\\
2 & 140 & $(1.8, 10.26)$& $(2.3, 6.07)$&$(3.4, 2.84)$ & $(6.6, 0.99)$\\
3 & 140 &$(1.8, 10.26)$&$(2.3, 6.07)$&$(3.4, 2.84)$&$(6.6, 0.99)$\\
4 & 70&$(1.8, 9.83)$&$(2.5, 6.03)$&$(3.2, 2.83)$&$(6.0, 1.0)$\\
5 & 70&$(1.8, 9.83)$&$(2.2, 6.03)$&$(3.2, 2.83)$&$(6.0, 1.0)$\\
6 & 10&$(2.1, 10.03)$&$(2.5, 6.0)$&$(3.6, 2.79)$&$(6.2, 1.0)$\\
7 & 10&$(2.1, 10.03)$&$(2.5, 6.0)$&$(3.6, 2.79)$&$(6.2, 1.0)$\\
8 & 4 &$(2.1, 10.03)$&$(2.5, 6.0)$&$(3.6, 2.79)$&$(6.2, 1.0)$\\
9 & 4 &$(2.1, 10.03)$&$(2.5, 6.0)$&$(3.6, 2.79)$&$(6.2, 1.0)$\\
10 & 2 &$(2.1, 10.03)$&$(2.5, 6.0)$&$(3.6, 2.79)$&$(6.2, 1.0)$\\
11 & 2&$(2.1, 10.03)$&$(2.5, 6.0)$&$(3.6, 2.79)$&$(6.2, 1.0)$\\
12 & 1&$(2.1, 10.03)$&$(2.5, 6.0)$&$(3.6, 2.79)$& $(6.2, 1.0)$\\
13 & 1&$(2.1, 10.03)$&$(2.5, 6.0)$&$(3.6, 2.79)$& $(6.2, 1.0)$\\
\midrule


14 & 27,325 & $(2.3, 9.15)$& $(2.8, 6.29)$&$(4.2, 2.91)$ & $(9.0, 0.99)$\\
15 & 26,463 & $(2.3, 9.22)$& $(2.7, 6.34)$&$(4.2, 2.93)$ & $(9.0, 0.99)$\\
16 & 27,259 &$(2.3, 9.15)$&$(2.8, 6.29)$&$(4.2, 2.91)$&$(9.0, 0.99)$\\
17 & 26,875 &$(2.3, 9.18)$&$(2.7, 6.32)$&$(4.2, 2.92)$&$(9.0, 0.99)$\\
18 & 26,344 &$(2.3, 9.23)$&$(2.7, 6.34)$&$(4.2, 2.93)$& $(9.0, 0.99)$\\

\midrule


19 & 1 & $(2.1, 10.03)$ &$(2.5, 6.0)$&$(3.6, 2.79)$&$(6.2, 1.0)$\\
20 & 1 & $(2.1, 10.03)$&$(2.5, 6.0)$&$(3.6, 2.79)$&$(6.2, 1.0)$ \\
21 & 1 & $(2.1, 10.03)$&$(2.5, 6.0)$&$(3.6, 2.79)$&$(6.2, 1.0)$\\
22 & 1 &$(2.1, 10.03)$&$(2.5, 6.0)$&$(3.6, 2.79)$&$(6.2, 1.0)$\\
23 & 1 &$(2.1, 10.03)$&$(2.5, 6.0)$&$(3.6, 2.79)$&$(6.2, 1.0)$\\

24 & 2 &$(2.1, 10.03)$&$(2.5, 6.0)$&$(3.6, 2.79)$&$(6.2, 1.0)$\\
25 & 2 &$(2.1, 10.03)$&$(2.5, 6.0)$&$(3.6, 2.79)$&$(6.2, 1.0)$\\
26 & 2 &$(2.1, 10.03)$&$(2.5, 6.0)$&$(3.6, 2.79)$&$(6.2, 1.0)$\\
27 & 2 &$(2.1, 10.03)$&$(2.5, 6.0)$&$(3.6, 2.79)$&$(6.2, 1.0)$\\
28 & 2 &$(2.1, 10.03)$&$(2.5, 6.0)$&$(3.6, 2.79)$&$(6.2, 1.0)$\\

29 & 4 &$(2.1, 10.03)$&$(2.5, 6.0)$&$(3.6, 2.79)$&$(6.2, 1.0)$\\
30 & 4 &$(2.1, 10.03)$&$(2.5, 6.0)$&$(3.6, 2.79)$&$(6.2, 1.0)$\\
31 & 4 &$(2.1, 10.03)$&$(2.5, 6.0)$&$(3.6, 2.79)$&$(6.2, 1.0)$\\
32 & 4 &$(2.1, 10.03)$&$(2.5, 6.0)$&$(3.6, 2.79)$&$(6.2, 1.0)$\\
33 & 4 &$(2.1, 10.03)$&$(2.5, 6.0)$&$(3.6, 2.79)$&$(6.2, 1.0)$\\

34 & 10 &$(2.1, 10.03)$&$(2.5, 6.0)$&$(3.6, 2.79)$&$(6.2, 1.0)$\\
35 & 10 &$(2.1, 10.03)$&$(2.5, 6.0)$&$(3.6, 2.79)$&$(6.2, 1.0)$\\

\bottomrule
\end{tabular}

\end{specialtable}

\clearpage

\end{paracol}

\reftitle{References}


\externalbibliography{yes}
\bibliography{bibliography/architectures_methods.bib, bibliography/attacks.bib, bibliography/databases.bib, bibliography/dp_basics.bib, bibliography/dp_ml.bib, bibliography/medical_images.bib, bibliography/overview.bib, bibliography/fl_basics.bib, bibliography/privacy_FL.bib, bibliography/xray.bib, bibliography/meta.bib, bibliography/web.bib}

\end{document}